%% file: bare_jrnl_new_sample4.tex
\documentclass[lettersize,journal]{IEEEtran}

\usepackage{amsmath,amsfonts}
\usepackage{array}
\usepackage[caption=false,font=normalsize,labelfont=sf,textfont=sf]{subfig}
\usepackage{textcomp}
\usepackage{stfloats}
\usepackage{url}
\usepackage{verbatim}
\usepackage{graphicx}
\usepackage{cite}
\usepackage{comment}
\usepackage{booktabs}
\usepackage{makecell}
\usepackage{multirow}
\usepackage[noend]{algpseudocode}
\usepackage[utf8]{inputenc}     
\usepackage{algorithm}
\usepackage{xcolor}

\begin{document}

\title{\fontsize{26pt}{28pt}\selectfont Split Federated Learning Architectures for \\ High-Accuracy and Low-Delay Model Training}

\author{Yiannis Papageorgiou, Yannis Thomas, Ramin Khalili, and Iordanis Koutsopoulos
\thanks{Y. Papageorgiou, Y. Thomas and I. Koutsopoulos are with the Department of Informatics, Athens University of Economics and Business, Athens, Greece. R. Khalili is with Huawei Technologies Duesseldorf Gmbh, Munich, Germany.}}

\maketitle

\begin{abstract}
Can we find a network architecture for ML model training so as to optimize training loss (and thus, accuracy) in Split Federated Learning (SFL)? 
And can this architecture also reduce training delay and communication overhead? 
While accuracy is not influenced by how we split the model in ordinary, state-of-the-art SFL, in this work we answer the questions above in the affirmative.
Recent Hierarchical SFL (HSFL) architectures adopt a three-tier training structure consisting of clients, (local) aggregators, and a central server.
In this architecture, the model is partitioned at two partitioning layers into three sub-models, which are executed across the three tiers.
Despite their merits, HSFL architectures overlook the impact of the partitioning layers and client-to-aggregator assignments on accuracy, delay, and overhead. 
This work explicitly captures the impact of the partitioning layers and client-to-aggregator assignments on accuracy, delay and overhead by formulating a joint optimization problem.
We prove that the problem is NP-hard and propose the first accuracy-aware heuristic algorithm that explicitly accounts for model accuracy, while remaining delay-efficient.
Simulation results on public datasets show that our approach can improve accuracy by 3\%, while reducing delay by 20\% and overhead by 50\%, compared to state-of-the-art SFL and HSFL schemes.
\end{abstract}

\input{intro}
\input{related}
\input{system}
\input{problem}
\input{algorithm}
\input{evaluation}
\input{conc}
%\begin{comment}
%\section*{Acknowledgement} This work was conducted in the context of the Horizon Europe project PRE-ACT (Prediction of Radiotherapy side effects using explainable AI for patient communication and treatment modification). It was supported by the European Commission through the Horizon Europe Program (Grant Agreement number 101057746), by the Swiss State Secretariat for Education, Research and Innovation (SERI) under contract number 22 00058, and by the UK government (Innovate UK application number 10061955).
%\end{comment}
\bibliographystyle{IEEEtran}
\bibliography{references}
\end{document}

%% file: intro.tex
\section{Introduction}
%The proliferation of devices that collect voluminous data has driven the adoption of collaborative distributed learning paradigms.
\emph{Split Federated Learning}~(SFL) ~\cite{thapa:2022} is an emerging Machine Learning~(ML) paradigm that combines the advantages of \emph{Federated Learning}~(FL)~\cite{mcmahan:2017} in terms of data privacy preservation with those of \emph{Split Learning}~(SL)~\cite{vepakomma:2018} on conserving resources for resource-constrained devices.
SFL combines the aggregation and offloading features of FL and SL respectively, thus enabling privacy-preserving training without high computational requirements at the clients.  
In FL, clients train local models on their private data and send the models to a server for aggregation into a global model.
SL offers computation offloading: the model is partitioned into two sub-models at a \emph{cut layer}, the first sub-model is trained at the clients and the second at the server, reducing the client-side computation.

SFL schemes suffer from long training delays due to two fundamental issues.
First, the \textit{backward locking effect} forces clients to wait for the server to complete its computations, thus remaining idle for long periods\cite{imteaj:2021}.
Second, client heterogeneity leads to the \textit{straggler effect}, where strong clients wait for weak clients to complete their computations, increasing training delay~\cite{ye:2023}.
To mitigate the former effect, \emph{local-loss learning} has been proposed, allowing clients to train in parallel with the server, using the local-loss computed at the cut layer and, therefore, reducing delay\cite{han:2021}.
The straggler effect has been addressed by several SFL schemes, through techniques such as asynchronous model updates~\cite{xu:2023}, helper clients~\cite{tirana:2024}, \cite{tirana:2024mp} and selection of personalized cut layers per client\cite{samikwa:2022}.

\begin{figure}
\label{fig:new_system}
\hspace{-0.01cm}\includegraphics[angle=0,trim={1.5cm 2.5cm 16.0cm 6.4cm},clip,scale=0.6]{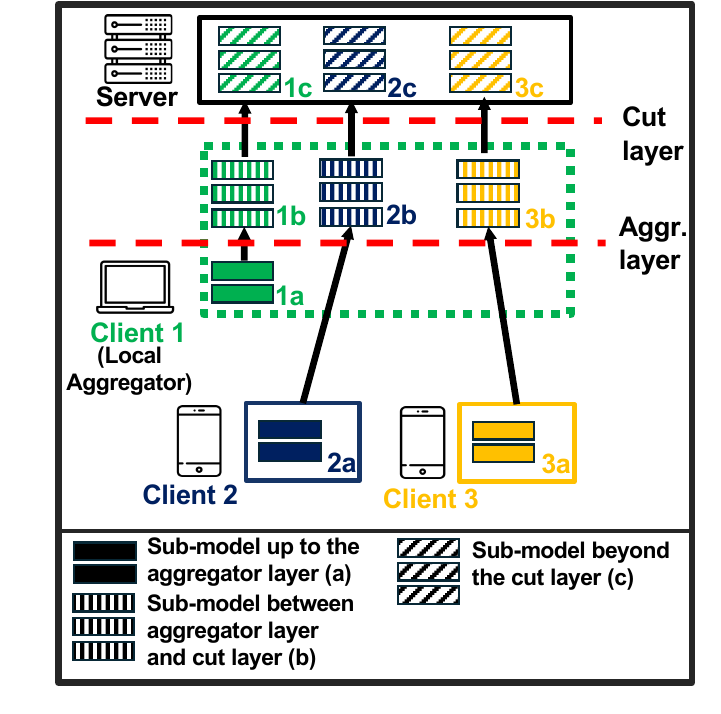}
\caption{Example network topology that presents a network architecture with three clients, where each client trains the model in collaboration with the server.
Client 1, being computationally stronger, is selected as a local aggregator.
The model is partitioned into three sub-models (a, b, and c) as defined by the aggregator layer and the cut layer.
The clients, local aggregators and server train the first (a), middle (b) and last (c) sub-model respectively.}
\label{fig:topology}
\end{figure}

Given that methods addressing the straggler effect are typically different from local-loss learning, they can be combined to address both issues~\cite{han:2021,shin:2023,mohammadabadi:2024,papageorgiou:2025}.
For instance, the hierarchical design introduced in~\cite{papageorgiou:2025} relies on partitioning the model into multiple parts to enhance parallelism and model aggregation frequency.
In detail, the model is partitioned into three sub-models, defined by two partitioning layers, the aggregator layer and the cut layer, which are executed in parallel across three different nodes.
The \emph{clients}, \emph{local aggregators} and \emph{server} train the first, middle and last sub-model respectively, as shown in Fig.~\ref{fig:topology}.
Local aggregators provide an additional aggregation level, enabling frequent model aggregation at lower communication overhead, which improves accuracy~\cite{lin:2024}, \cite{papageorgiou:2025}.

\begin{figure}
\begin{tabular}{c c}
\hspace{-0.2cm}\includegraphics[angle=0,trim={0cm 0cm 0cm 0cm},clip,scale=0.31]{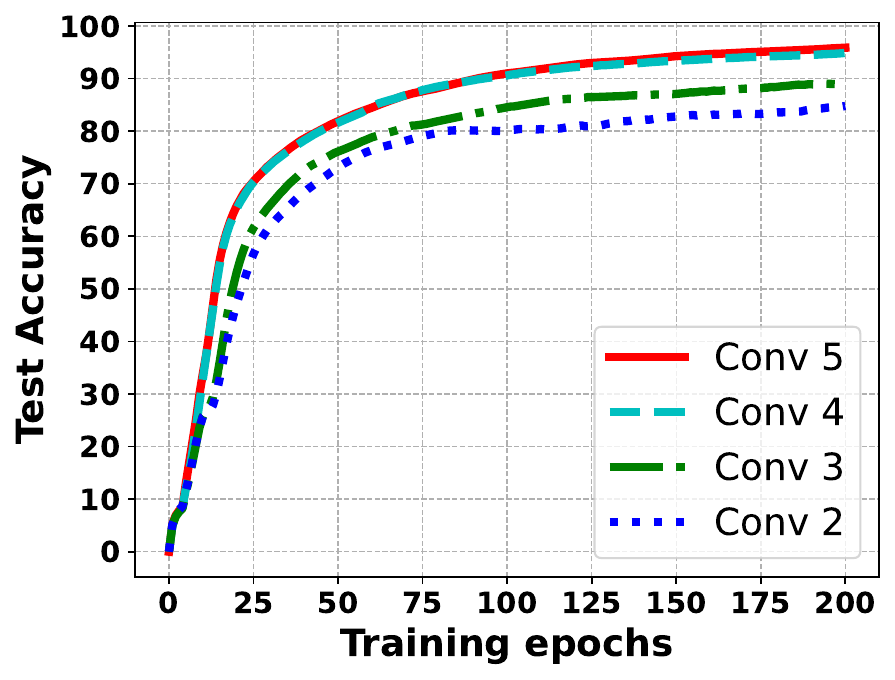} & \hspace{-0.3cm}\includegraphics[angle=0,trim={2cm 0cm 0cm 0cm},clip,scale=0.31]{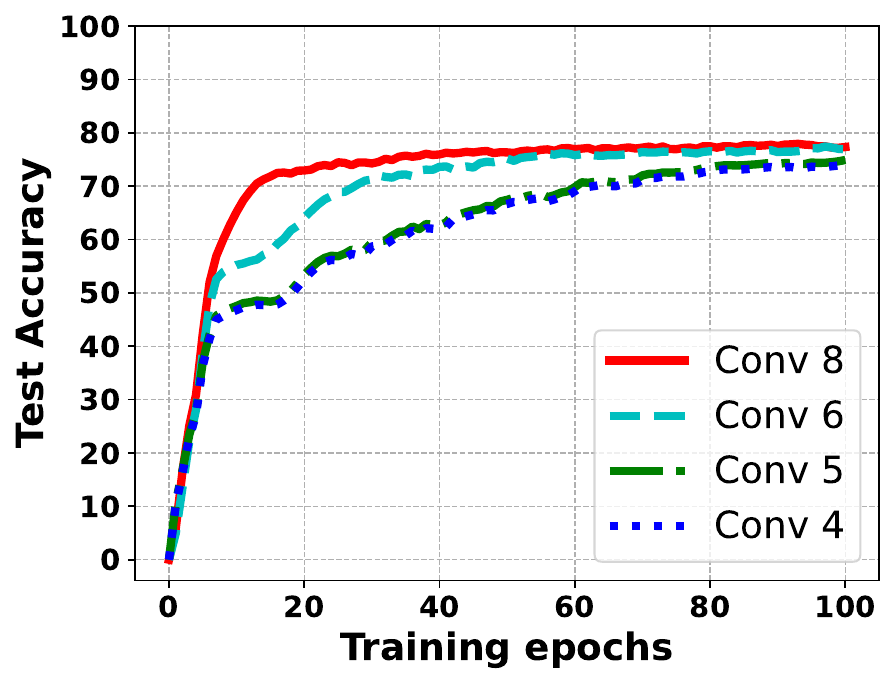} \\
\hspace{-1.1cm} \footnotesize (a) Alexnet & \hspace{-0.0cm} \footnotesize (b) VGG-11 \\
\end{tabular}
\caption{Test accuracy versus training epochs for different cut layers during the training of AlexNet and VGG-11 models across 100 clients with local-loss learning.}
\label{fig:motivation}
\vspace{-0.3cm}
\end{figure}

These hierarchical designs expand the design space of efficient algorithms in order to influence key performance metrics such as accuracy, delay and overhead, by adjusting the selection of different cut layers and the assignment of clients to local aggregators.
%introduce a new set of problems, such as the selection of multiple cut layers and the assignment of clients to aggregators.
However, existing HSFL schemes largely overlook this enhanced design space~\cite{han:2021,shin:2023,mohammadabadi:2024,papageorgiou:2025}.
For example, a common assumption underlying existing SFL schemes is that accuracy is invariant to the selection of the cut layer, and that the cut layer selection mainly affects delay and overhead.
In this work, we provide evidence that the selection of a suboptimal cut layer can severely degrade the accuracy.
Fig.~\ref{fig:motivation} presents the test accuracy of the AlexNet and VGG-11 models in time (training epochs) for different cut layer selections.
The results reveal that selecting convolution layer~2 as the cut layer results in lower accuracy, whereas selecting convolution layer~5 achieves higher accuracy.
Therefore, we emphasize the need to steer attention towards accuracy-aware cut layer selection (and SFL operation in general). 

To the best of our knowledge, this work is the first to consider a HSFL architecture with clients, aggregators, and a server, in which the partitioning layers at the clients and the aggregator as well as the client-to-aggregator assignments are jointly selected.
We show that this joint selection can improve the model accuracy, but also reduce the training delay and the overhead.
Specifically, the system must jointly select the partitioning layers and client-to-aggregator assignments, so as to minimize delay and satisfy accuracy constraints.

The joint selection of the partitioning layers and client-to-aggregator assignments constitutes a hard problem.
Therefore, we propose an accuracy-aware heuristic algorithm, named \textit{Accuracy-Aware Hierarchical Federated Learning with Local Loss}~(AA HSFL-ll) that caters to accuracy, while being also delay-efficient.
AA-HSFL operates in two phases: first, it finds the set of cut layers that achieve high accuracy and, second, uses this set to jointly select the partitioning layers and client-to-aggregator assignments that minimize the training round delay, i.e., the end-to-end delay of one round.

The contributions of this paper are summarized as follows:
\begin{itemize}
    \item We formulate a joint optimization problem over partitioning layers' selection and client-to-aggregator assignments that aims to minimize training delay with accuracy constraints and prove its NP-hardness.
    
    \item We propose an accuracy-aware algorithm that jointly optimizes the selection of partitioning layers and client-to-aggregator assignments, achieving delay and communication overhead reduction, while increasing model accuracy.
    
    \item We demonstrate that our algorithm can improve accuracy by 3\%, while reducing delay by 20\% and overhead by 50\%, compared to existing SFL and HSFL schemes.
    
    \item We also show that our algorithm can achieve a near-optimal solution with respect to the optimal solution computed via exhaustive search, while preserving low computational complexity and robustness to system changes, including additional client-side processing tasks and fluctuations in network transmission rates.
\end{itemize}

The rest of the paper is organized as follows. Section \ref{sec:relatedwork} reviews related work, and Section \ref{sec:system} presents the system architecture. Section \ref{sec:problem} formulates the problem and proves its NP-hardness, while Section \ref{sec:algorithm} proposes a heuristic solution. Section \ref{sec:evaluation} presents numerical results, and Section \ref{sec:conclusion} concludes the paper.

%% file: related.tex
\section{Related Work}\label{sec:relatedwork}
\subsection{Related work}
\textbf{Split Federated Learning (SFL):} 
Split Federated Learning (SFL) integrates Federated Learning (FL) and Split Learning (SL) to enable privacy-preserving model training while alleviating the limited computational capabilities of clients \cite{thapa:2022}.
%Split Federated Learning (SFL) combines the advantages of Federated Learning (FL) in terms of data privacy preservation with those of Split Learning (SL) on conserving resources for resource-constrained devices\cite{thapa:2022}.
In SFL, the global model is partitioned at the cut layer into the client-side and the server-side models, which are executed at the clients and the server, respectively, without compromising model accuracy \cite{vepakomma:2018}.
By parallelizing client-side training across clients, SFL can significantly reduce the training delay.

However, SFL still suffers from the \textit{backward locking effect}, where clients must wait for gradients computed at the server before proceeding with BackPropagation (BP), as well as the \textit{straggler effect}, where stronger clients wait for weaker ones to complete their computations, thereby increasing delay.
Existing studies address these effects, through model partitioning and aggregation optimization, either independently or jointly.
%Finish the introduction of SFL

\textbf{Mitigating the straggler effect in SFL:}
Kim et al. \cite{kim:2023} propose a cut layer selection strategy that minimizes a weighted cost function incorporating delay, energy consumption, and communication overhead.
Similarly, \cite{samikwa:2022} and \cite{wu:2023} develop topology-aware cut layer selection strategies that minimize delay by accounting for network conditions.
%Lin et al. \cite{lin:2024efficient} propose an efficient SFL scheme that jointly optimizes model partitioning, by determining a personalized cut layer per client, and transmission rate management, introducing a tunable last layer gradient aggregation interval to reduce training delay, albeit at the cost of degraded accuracy due to aggregation of models of different layers.
Lin et al. \cite{lin:2024efficient} propose an SFL scheme that jointly optimizes model partitioning by selecting a personalized cut layer per client and managing transmission rates, while introducing a tunable gradient aggregation interval that reduces delay, but degrades accuracy due to aggregation of models of different layers.

A long thread of papers explores the use of helpers, i.e., computationally stronger clients that assist weaker clients in executing client-side training.
Wang et al. \cite{wang:2023} and Yao et al. \cite{yao:2025} jointly determine the cut layer and client-to-helper assignments based on clients’ computational capabilities.
Tirana et al. \cite{tirana:2024} extend this approach by considering client memory limitations, while \cite{tirana:2024mp} further generalizes it by allowing multiple helpers per client.
The same authors extend their work by formulating a multiobjective optimization problem that studies the trade-off between energy consumption and delay \cite{tirana:2025}.

%There are some recent works on the optimization of Hierarchical Split Federated Learning (HSFL) schemes \cite{Tengxi:2022},\cite{Khan:2024}, \cite{Zheng:2025}, which provide the merit of the hierarchical computation and aggregation.
%Some define the term hierarchical specifically in the context of model aggregation (MA) \cite{Tengxi:2022}, \cite{Khan:2024}, or computation \cite{tirana:2024}, while others adopt a more comprehensive approach that also includes optimizing cut layer selection and MA interval configuration\cite{Zheng:2025}.
There are recent works on the optimization of Hierarchical SFL (HSFL) schemes \cite{Tengxi:2022}, \cite{Khan:2024}, \cite{Zheng:2025}, which exploit the benefits of hierarchical computation and aggregation.
Some studies define the term \emph{hierarchical} in the context of model aggregation (MA) \cite{Tengxi:2022}, \cite{Khan:2024}, or hierarchical computation \cite{tirana:2024}, while others adopt an approach that includes optimizing cut layer selection and MA interval \cite{Zheng:2025}.

\textbf{Mitigating the backward locking effect in SFL:}
Another line of work focuses on local-loss learning, where client-side training proceeds without waiting for gradients computed at the server, enabling parallel client-side and server-side BP \cite{nokland:2019,han:2021,Seungeun:2022}.
This parallel execution effectively mitigates the backward locking effect and reduces delay; however, it comes at the cost of degraded accuracy as client-side updates rely on gradients computed at the cut layer rather than on gradients at the last layer\cite{nokland:2019}.
These approaches are summarized in Table~\ref{tab:relatedworks} under the Local-loss column.

\textbf{Mitigating both effects in SFL:}
Recent works combine local-loss learning with personalized cut layer selection, based on clients' computation throughput, to mitigate both backward locking and straggler effects in heterogeneous environments \cite{shin:2023},\cite{mohammadabadi:2024}.
While these approaches reduce delay, they also report non-negligible degradation in accuracy, indicating that delay-oriented model partitioning can harm accuracy.
This accuracy degradation arises since the selection of the cut layer determines the local-loss gradients used during client-side BP.

%Our prior work \cite{papageorgiou:2025} introduced a HSFL architecture, termed HSFL-ll, that combines local-loss learning with helper clients by introducing an additional partitioning layer, referred to as aggregator layer, between clients and helpers, referred to as local aggregators.
Our prior work \cite{papageorgiou:2025} introduced a HSFL architecture, termed HSFL-ll, that combines local-loss learning with helper clients.
The architecture introduces an additional partitioning layer, called aggregator layer, between clients and helper clients acting as local aggregators. 
%enabling a three-tier execution of the model across clients, aggregators, and the central server.
However, in \cite{papageorgiou:2025}, the selection of the partitioning layers and the client-to-aggregator assignments, were fixed, and their impact on accuracy, delay and overhead was not explored.
To the best of our knowledge, this work is the first to study the impact of partitioning layers selection and client-to-aggregator assignments, and to propose an accuracy-aware algorithm that jointly optimizes these design variables to improve accuracy, while reducing delay and overhead.
%A detailed comparison with existing approaches is provided in Table~\ref{tab:relatedworks}, which indicates the paper's contribution.
%shows that the proposed solution is the first to jointly optimize these design choices in an accuracy-aware manner.
\begin{table*}[t]
\centering
\caption{Comparison of existing SFL schemes and the proposed accuracy-aware solution.}
\label{tab:relatedworks}
\begin{tabular}{|l|c|c|c|c|c|c|}
\hline
\multirow{2}{*}{Reference} 
& \multicolumn{4}{c|}{Straggler effect} 
& \multicolumn{1}{c|}{Backward locking effect} 
& Accuracy-awareness \\
\cline{2-7}
& Topology-awareness 
& Helpers
& Cut layer selection
& Aggregation strategy
& Local-loss 
& Joint optimization \\
\hline

\cite{samikwa:2022}, \cite{wu:2023}, \cite{kim:2023}
& \checkmark &  & \checkmark &  &  &  \\
\hline

\cite{lin:2024}, \cite{lin:2024efficient}
& \checkmark &  & \checkmark & \checkmark &  &  \\
\hline

\cite{han:2021}, \cite{nokland:2019}, \cite{Seungeun:2022}
&  &  &  &  & \checkmark &  \\
\hline

\cite{wang:2023}, \cite{yao:2025}
&  & \checkmark & \checkmark &  &  &  \\
\hline

\cite{tirana:2024}, \cite{tirana:2024mp}
&  & \checkmark & \checkmark &  &  &  \\
\hline

\cite{wu:2025}
& \checkmark & \checkmark & \checkmark &  &  &  \\
\hline

\cite{Tengxi:2022}, \cite{Khan:2024}, \cite{Zheng:2025}
& \checkmark & \checkmark & \checkmark & \checkmark &  &  \\
\hline

\cite{shin:2023}, \cite{mohammadabadi:2024}
&  &  & \checkmark &  & \checkmark &  \\
\hline
Prior work \cite{papageorgiou:2025}
& \checkmark & \checkmark & \checkmark & \checkmark & \checkmark &  \\
\hline
\textbf{Proposed}
& \checkmark & \checkmark & \checkmark & \checkmark & \checkmark & \checkmark \\
\hline
\multicolumn{7}{l}{\footnotesize Aggregation strategy includes models' aggregation location and frequency (MA interval).}
\end{tabular}
\end{table*}

%% file: system.tex
\section{Accuracy-Aware Hierarchical Split Federated Learning with Local Loss}\label{sec:system}
\begin{figure}
\label{fig:new_system}
\hspace{-0.01cm}\includegraphics[angle=0,trim={1.5cm 2.5cm 8.5cm 2.5cm},clip,scale=0.45]{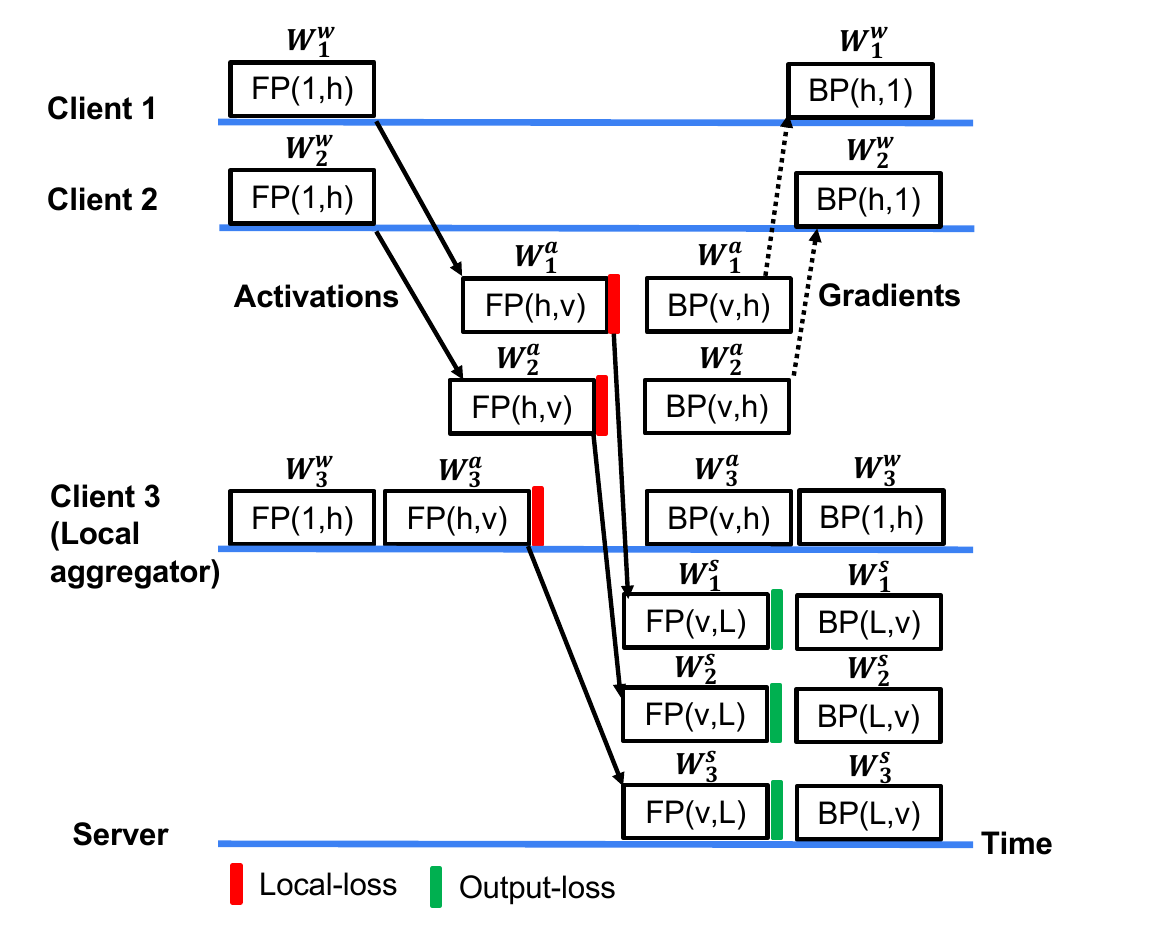}
\caption{Batch processing with one server and 3 clients, where client 3 acts as local aggregator. The figure depicts the training tasks (rectangles) executed by each client in time (x-axis). The horizontal blue lines mark the execution zone of each node; tasks within this zone are executed by the corresponding node; multi-threaded execution is implied when multiple tasks in the same zone overlap in time. For each task, we present the trained sub-model, the training ``direction'' (FP or BP) and the trained layers, e.g., $W^w_3$ $FP(1,h)$ encodes that FP of layers 1 to h in the weak-side sub-model of client 3. The arrows symbolize the network transmission of activations and gradients in the training chain.
%Each client $n$ performs forward propagation (FP) on its weak-side model $W^w_n$ and transmits the activations of the aggregator-layer $h$ to its assigned local aggregator. Local aggregator continue FP on the aggregator-side model $W^a_n$ up to the cut layer $v$ and transmit the activations to the server.
%Using the local-loss at the cut layer (red rectangle), backpropagation (BP) is then performed on the aggregator-side and weak-side models, while the server simultaneously executes FP and BP on the server-side models using the output layer's loss (green rectangle).
}
\label{fig:batch_workflow_new}
\end{figure}
\subsection{System design}

%In this paper, and without loss of generality, we consider the work in~\cite{papageorgiou:2025} as the underlying SFL solution incorporating the proposed algorithm described in Section \ref{sec:algorithm}.
In the following, we provide a technical description of the network architecture and the processes executed during the training procedure.
%Hereafter, we consistently use \emph{FP} and \emph{BP} to denote forward and back propagation, respectively.

\subsubsection{Network}
The network consists of three node types: one \textbf{server}, $N$ \textbf{clients}, a subset of which act as \textbf{local aggregators}.
The nodes are interconnected via wireless communication links, forming a fully connected network graph.
The computation throughput of client $n$ is denoted by $p_n$ (in FLOPS/sec), while $p_s$ refers to computation throughput of the server. 
The transmission rate of the link between any two nodes $i$ and $j$ is represented by $r_{i,j}$ (in bytes/sec). 
We assume that the computation throughput of nodes, the transmission rates and the network topology remain stable.
The server collects the required system information and orchestrates the training process. %using the proposed algorithm in Section \ref{sec:algorithm}.
The fraction of clients operating as local aggregators is denoted by $\lambda$, where $\lambda \in (0,1)$, and each client can be assigned only to one local aggregator.
\subsubsection{Model partitioning}
The model is denoted with $W$ and consists of $L$ sequential layers. A client $n$ has an individual instance on the model, $W_n$, trained on its own individual dataset, $D_n$.
The model is partitioned at two layers, namely, the aggregator layer $h$, and cut layer $v$, thus delivering three complementary sub-models, namely, weak-side model $W^w_n$, consisting of layers $\{1, \dots, h\}$, aggregator-side model $W^a_n$, consisting of layers $\{h+1, \dots, v\}$, and server-side model $W^s_n$, consisting of layers $\{v+1, \dots, L\}$. 
These sub-models are trained at the client, the assigned aggregator and the server, respectively.

\subsubsection{Training} Training is based on \emph{batch processing} and \emph{local \& global aggregation} procedures.
Each client performs batch processing, which is training on multiple data samples (\emph{batch}) at once to reduce the overall delay.
Then, the individual models are averaged via the aggregation procedures at either the local aggregators or the server.
The overview of the training procedure is outlined in the following.
%as well as depicted in Fig.~\ref{fig:local_aggregation}.

\paragraph{Batch processing} 
We outline the batch processing steps for 3 clients of a model with $L$ layers, where client 3 acts as local aggregator, as depicted in Fig.~\ref{fig:batch_workflow_new}.
Client $n$ performs Forward Propagation~(FP) on its weak-side model, $W^w_n$, until the aggregator layer $h$, then, transmits the output of the layer $h$ activation function (\emph{activations}) to its assigned local aggregator.
%The local aggregator $k$ performs FP on the aggregator-side models $W^a_n$ for each client $n$ assigned and, then, transmits the corresponding activations to the server.
The local aggregator $k$ performs FP on the aggregator-side model $W^a_n$ of each assigned client $n$, and then transmits the corresponding activations to the server.
Without waiting for server response, the local aggregator computes the local-loss (Local-loss in Fig. \ref{fig:batch_workflow_new}) error signals (\emph{gradients}), at the cut layer $v$ and performs Back Propagation~(BP) on the aggregator-side models.
Finally, the local aggregator transmits the layer $h$ gradients back to the corresponding clients, allowing them to complete BP on their weak-side models.

The server, in parallel, upon receiving the cut layer activations from the local aggregator(s), performs FP on the server-side models, computes the loss at the final layer (Output-loss in Fig. \ref{fig:batch_workflow_new}), and performs BP on the respective server-side models.

\paragraph{Local \& Global aggregation}
The batch processing procedure is repeated by all clients for all batches in the dataset, this is called a training \emph{epoch}.
A number of epochs $E$ defines a training \emph{round}. 
This distinction is important when studying the model aggregation periods: the \emph{local} aggregation at the local aggregator and server occurs every epoch, while the \emph{global} aggregation at the server reoccurs every round. 

In the local aggregation, each local aggregator aggregates the aggregator-side models received from its assigned clients, using the FedAvg algorithm\cite{mcmahan:2017}; no models or gradients are sent over the network.
Also, server aggregates the server-side models of all clients.
In the global aggregation, the server receives the aggregator-side models (from the aggregators) and the weak-side models (from the clients).
Then, it computes the aggregated aggregator-side model and aggregated weak-side model using the FedAvg algorithm \cite{collins:2022}.
These updated models are subsequently sent back to be used in the next round.

Repeating the local aggregation every epoch boosts accuracy without excessive communication overhead.
Accuracy is improved by the frequent aggregations of the aggregator-side models \cite{lin:2024},\cite{papageorgiou:2025}, with lower overhead in our case.
In contrast, the resource-expensive global aggregation is conducted on a longer time interval to keep overhead within a practical level.   
\begin{table}
\caption{Notation Table}
\label{tab:notation_table}
\centering
\begin{tabular}{|l|p{0.7\linewidth}|} % Adjust 0.7\linewidth as needed
\hline
$a_l$ & Weights of layer $l$ (bits) \\
\hline
$B$ & Size of data batch \\
\hline
$Q$ & Number of batch processing executions within a epoch \\
\hline
$g_l$ & Activations/gradients of layer $l$ (bits) \\
\hline
$E$ & Number of epochs \\
\hline
$T_i$ & Training delay of phase $i, i \in [1, 2, 3]$\\
\hline
$f_l$ & Computation workload of layer $l$ (FLOPS)\\
\hline
$h$, $v$ & Aggregator layer, Cut layer\\
\hline
$\mathcal{N}, \mathcal{K}$ & Set of clients, Set of local aggregators \\
\hline
$D_n$ & Dataset of client $n$ \\
\hline
$p_n$ & computation throughput of client $n$ (FLOPS/sec)\\
\hline
$p_s$ & computation throughput of server (FLOPS/sec)\\
\hline
$r_{i,j}$ & Transmission rate of link between clients $i$ and $j$ (bytes/ sec)\\
\hline
$acc(v, E')$ & Achieved model accuracy with cut layer $v$ after $E'$ epochs\\
\hline
$\lambda$ & Fraction of clients operating as local aggregators\\
\hline
$\gamma$ & Client heterogeneity ratio\\
\hline
$W^w_n$ & Weak-side model of client $n$ \\
\hline
$W^a_{n}$ & Aggregator-side model of client $n$  \\
\hline
$W^s_n$ & Server-side model for client $n$\\
\hline
$x_{n,k,l}$ & Binary variable indicating whether layer $l$ of client's $n$ model is assigned to local aggregator $k$\\
\hline
$\mathcal{X}$ & Set of binary variables $x_{n,k,l}$\\
\hline
\end{tabular}
\end{table}
\subsection{Training delay analysis}
To quantify the training delay we need to assess the execution delay of one training round. We decompose the round into three consecutive phases and denote their corresponding delay with $T_i, i \in [1, 2, 3]$.
Phases $1$ and $3$ involve downloading the client’s model from the server and uploading it back, respectively.
Phase $2$ implements a number of batch processing executions $Q$, derived by the dataset's and batch data size.
In detail, this number is $Q=\lceil{D_n}/{B}\rceil, \forall n \in \mathcal{N}$, where $B$ is the batch data size and $D_n$ is the dataset of client $n$.

Regarding computational cost, let $a_l$ denote the weights of layer $l$ (bits), let $g_l$ denote the activations/gradients of layer $l$, and let $f_l$ denote the corresponding computational workload (FLOPS) for FP of layer $l$.
The computational workload (FLOPS) for BP of the same layer $l$, is approximately twice the $f_l$ \cite{FLOPS}.
\footnote{These computational cost parameters can be also measured offline, other than using standard formulas.}
Let also $h$ denote the aggregator layer and $v$ the cut layer.
We further denote by $\mathcal{K}$ the set of local aggregators. 
Also, the binary assignment variable $x_{n,k,l}$ indicates whether layer $l$ of client's $n$ model is assigned to local aggregator $k$ and $\mathcal{X}$ the set of these variables.
\begin{equation}\label{eqn:binary_variable}
    x_{n,k,l} \in \{0,1\}, \quad \forall i, k \in \mathcal{N}.
\end{equation}
It also holds that each client $n$ is assigned to a single aggregator (client) $k$:
\begin{equation}
\label{eqn:single_aggregator}
\sum_{k \in \mathcal{N}} x_{n,k,h+1} \leq 1, \quad \forall n \in \mathcal{N}
\end{equation}
Also, the layers $h+1 \dots v$ of the model of client $n$ should be trained on the same local aggregator $k$.
\begin{equation}
\label{eqn:h_v_layers}
x_{n,k,l} = x_{n,k,h+1}, \quad \forall n, k \in \mathcal{N}, \; \forall l \in  [h+1,v]
\end{equation}
% The round includes computations that occur at different time intervals.
% Specifically, the batch workflow shown in Fig. \ref{fig:local_aggregation} is repeated per data batch, while the local and global aggregations are performed once per epoch and once per round respectively.
% The aggregation of client-side (or server-side) models using the FedAvg\cite{mcmahan:2017} has low computational complexity, leading to negligible delay, and therefore is excluded from the analysis.

During phase 1, the server broadcasts the weak-side model $W^w$ to the clients and both aggregator-side model $W^a$ and $W^w$ to the aggregators. The delay $T_1$ is the maximum download delay across all recipients and transmitted models:
\begin{equation}\label{eqn:d0}
T_1 = \max \Big\{  
    \max_{n \in \mathcal{N}} \Big\{  
        \displaystyle\frac{\sum\limits_{l=1}^{h} a_l}{r_{s,n}} 
    \Big\}  , \quad
    \max_{k \in \mathcal{K}} \Big\{  
        \displaystyle\frac{\sum\limits_{l=1}^{v} a_l}{r_{s,k}} 
   \Big\}  
\Big\} 
\end{equation}

During phase $2$, the batch processing procedure is repeated $Q$ times within an epoch. Delay of phase 2, $T_2$, is the sum of the following delays of the slowest client-aggregator pair:
\begin{itemize}
\item \textbf{FP delay of weak-side models:}
Each client performs FP on its weak-side model, using its local data.
This delay corresponds to the fraction of the weak-side model's FP computational workload and the client's $n$ computation throughput $p_n$, ($ \sum\limits_{l=1}^{h}\cdot f_l/ p_n $).

\item \textbf{Transmission delay of aggregator layer activations:}
Each client $n$ transmits to their local aggregator $k$ the aggregator layer activations.
The delay is equal to the transferred data's size $g_h$ divided by the link transmission rate $r_{n,k}$, ($g_{h}/r_{n,k}$).  

\item \textbf{FP delay of aggregator-side models:}
Each local aggregator $k$ executes FP on the aggregator-side models on behalf of its assigned clients.
This delay corresponds to the computational workload of the number of aggregator-side models' for FP divided by the aggregator's computation throughput $p_k$. Thus, the delay for aggregator $k$ is: ($
(\sum\limits_{m \in \mathcal{N}} \sum\limits_{l=h+1}^{v} x_{m,k,l} \cdot f_l )/p_k$).

\item \textbf{Transmission delay of cut layer activations:}
After completing the aggregator-side models' FP, the local aggregators transmit to the server the cut layer activations $g_v$ for each assigned client.
The delay is equal to the size of the transferred data $g_v$ divided by the link transmission rate $r_{n,k}$, ($g_{v}/r_{k,s}$). 
The total delay of these four steps is denoted with $T_{FP}$ and equals to:
\begin{align}
\label{eqn:d1}
T_{FP} =
\max_{n \in \mathcal{N}}
\Big\{
&\frac{\sum\limits_{l=1}^{h} f_l}{p_n}
+ \sum\limits_{k \in \mathcal{K}} x_{n,k,h+1} \frac{g_h}{r_{n,k}}
\nonumber \\
&+ \sum\limits_{k \in \mathcal{K}} x_{n,k,h+1}
\frac{
\sum\limits_{m \in \mathcal{N}}
\sum\limits_{l=h+1}^{v} x_{m,k,l}  f_l
}{p_k}
\nonumber \\
&+ \sum\limits_{k \in \mathcal{K}} x_{n,k,v+1} \frac{g_v}{r_{k,s}}
\Big\}.
\end{align}

\item \textbf{FP and BP delay of server-side models :}
This step involves the execution of the server-side models' FP and BP using the collected activations from all clients from the local aggregators.
Its delay $T_{s}$ equals with the total computational workload for executing FP and BP on all server-side models divided by the server's computation throughput $p_s$. Using the approximation that the BP workload is twice the FP workload, we would have:
\begin{equation}
\label{eqn:server_delay}
    T_{s}=\frac{3 \sum\limits_{n =1}^{N} \sum\limits_{l = v+1}^{L} f_l }{p_s}
\end{equation}
\item \textbf{BP delay of aggregator-side models:}
The Aggregator-side models' BP delay is defined analogously to the FP delay, yet the computational workload of layers is double, and corresponds to the total BP computational workload of the assigned aggregator-side models divided by the $p_k$.

\item \textbf{Transmission delay of aggregator layer gradients:} 
Upon completing the BP on the aggregator-side model, the local aggregator $k$ sends the layer $h$ gradients to its assigned clients with the respective transmission delay.

\item \textbf{BP delay of weak-side model:}
Then, the clients execute BP on their weak-side model after receiving the aggregator layer $h$ gradients and the respective delay is analogical to the weak-side model's FP delay.

The delay of these four steps, denoted by $T_{BP}$, equals the maximum of the server-side models' FP \& BP delay $T_s$ and the total delay of the remaining three steps:
\begin{align}
\label{eqn:d2}
T_{BP}=\max\Big\{ &T_{s},\nonumber \\
& \max_{n \in \mathcal{N}} \Big\{  
    \sum\limits_{k \in \mathcal{K}} x_{n,k,h+1}
\frac{
2 \sum\limits_{m \in \mathcal{N}}
\sum\limits_{l=h+1}^{v} x_{m,k,l}  f_l
}{p_k}
    \nonumber \\
&
    + \sum\limits_{k \in \mathcal{K}} x_{n,k,h+1} \frac{g_h}{r_{k,n}}
 \quad + \displaystyle\frac{2 \cdot \sum\limits_{l = 1}^{h} f_l}{p_n}
\Big\}  
\Big\}  
\end{align}
\end{itemize}
Therefore, the delay of phase 2, $T_2$, is equal to:
\begin{equation}
\label{eqn:t2}
T_2=T_{FP} + T_{BP}
\end{equation}
During phase $3$, each local aggregator $k \in \mathcal{K}$ uploads its aggregator-side and weak-side models to the server, and each client $n \in \mathcal{N/K}$ uploads its weak-side model to the server.
Since we assume symmetric wireless links (see Section \ref{sec:system}.A), $T_3$ is equal to $T_1$. Therefore:
\begin{equation}
\label{eqn:d4}
    T_3=T_1
\end{equation}

Taking into account (\ref{eqn:d0})--(\ref{eqn:t2}) and considering that phase 2 is executed $E \cdot Q$ times per round (i.e., $Q$ batch executions per epoch over $E$ epochs), the  round delay is given by:
\begin{equation}\label{eqn:dtotal}
T_{round}= T_1 + E \cdot Q \cdot T_2 + T_3
\end{equation}
Notably, the round delay depends on the joint selection of the aggregator layer $h$, the cut layer $v$, and the client-to-aggregator assignments $\mathcal{X}$.

%% file: problem.tex
\subsection{Problem Formulation -- Joint selection of partitioning layers \& client-to-aggregator assignments}\label{sec:problem}
We now formulate the problem of jointly selecting the partitioning layers, the cut and the aggregator layer, and the client-to-aggregator assignments.
Its objective is to minimize the training delay, while satisfying model accuracy constraints.

Let $acc(l,E')$ denote the achieved model accuracy when layer $l$ is selected as the cut layer after $E'$ epochs of training and $\max\limits_{l \in [2, \ldots , L-1]} \text{acc}(l,E')$ the maximum observed accuracy.
We focus on the selection of cut layer $v$, since it determines the local-loss gradients used during the aggregator-side and weak-side models' BP, and thus the resulting achieved accuracy.
%We consider that
Define $\mathcal{V}^*$ as the set of candidate cut layers whose achieved accuracy is within a tolerance threshold $thr$ of the maximum observed accuracy:
\begin{equation}
\mathcal{V}^* = \Big\{  v \mid \text{acc}(v,e) \geq \max_{l \in [2, \ldots , L-1]} \text{acc}(l,e) - thr \Big\} 
\label{eqn:candidate_cut_layers}
\end{equation}

The selected cut layer $v$ must belong to this set, i.e., $v \in \mathcal{V}^*$ (\ref{eqn:h_v_layers}).
The aggregator layer $h$ should be greater than the first layer and smaller than the cut layer $v$:
%\begin{equation}
%\label{eqn:aggregator_layer}
$1< h < v$ (16).
%\end{equation}
The cut layer $v$ should be greater than the second layer and smaller than the last layer $V$:
%\begin{equation}
%\label{eqn:cut_layer2}
$2< v < L$ (17).
%\end{equation}
Formally, we seek to solve the following optimization problem:
\begin{equation}\label{eqn:delay_opt_problem}
\begin{aligned}
&  \min_{h,v,\mathcal{X}} \quad T_{round} \\
& \text{subject to: } 
(\ref{eqn:binary_variable}),(\ref{eqn:single_aggregator}),(\ref{eqn:h_v_layers}), (\ref{eqn:candidate_cut_layers})
\end{aligned}
\end{equation}

%\subsection{Complexity of exhaustive enumeration}
Regarding the aggregator \& cut layer selection, an exhaustive method that evaluates the pairs of $(h, v)$ that minimizes delay $T_{round}$  (\ref{eqn:delay_opt_problem}) is the following. 
For $L$ layers, there are $L-2$ possible solutions for the aggregator layer $h$, i.e., after layers $1, 2, \ldots L-2$.
Say that $h=j$, there are $(L-1) - (j+1) +1$ solutions for the cut layer $v$, in the worst case that $V^*=L$, because it has to be after the $h$ layer and before the $L-1$ layer.
Thus, the number of possible $(h,v)$ combinations is:
$\sum_{h=2}^{L-1} \sum_{v=h+1}^{L-1} 1$ which is $O(L^2)$. 

Regarding the computation of client-to-aggregator assignments, for each $(h,v)$ pair, we need to explore every possible client-to-aggregator assignment.
With $K$ potential aggregators (up to $N$), the number of assignments grows exponentially with $N$.
%Say that there are $K$ potential aggregators, which can be up to $N$, the total number of possible assignments grows exponentially with $N$.
Thus, the overall search space complexity becomes $O(N^N*L^2)$ making exhaustive search infeasible for large $N$.
% For completeness, we also quantify the total communication overhead, which is used in the evaluation Section \ref{sec:evaluation}, incurred during a round, as a function of the selected partitioning layers and client-to-aggregator assignments:
% \begin{equation}\label{eq:overhead}
% CO(h,v,\mathcal{X})= 
% \sum_{i=1}^{N} \sum_{j=1}^{N} \sum_{l=1}^{v}
% x_{i,j,l} \cdot
% \begin{cases}
% 2 g_h Q, & \text{if } l = h \\
% 2 g_v Q, & \text{if } l = v \\
% 2 g_l,   & \text{otherwise}
% \end{cases}
% \end{equation}

\subsection{Problem's NP-hardness}
Joint problems of assignment and scheduling decisions, such as the facility location problem, are NP-hard, see, e.g.\cite{lawler:1993},\cite{lenstra:1990},\cite{vazirani:2001},\cite{cornuejols:1983}.
While one could in principle rely on exhaustive search methods \cite{nievergelt:2000e} to obtain the optimal solution, our experiments show that such approaches incur prohibitive computational overhead even for small problem instances.
Moreover, even when the aggregator and cut layers are fixed, the computation of client-to-aggregator assignments reduces to an instance of the single-allocation Restricted Facility Location Problem (RFLP)~\cite{Gimadi:2021}.
We search for the set of facilities (local aggregators) that minimize the cost of serving the demand (round delay) of clients (weak clients) that are uniquely associated with them (single allocation) under link capacity constraints (single hop communication between clients and local aggregators).
RFLP is known to be NP-hard \cite{Gimadi:2021}, hence the problem hardness is intuitive, even if a formal proof, e.g., through reduction from a known hard problem, is not straightforward. 
Thus, we propose a heuristic algorithm in the next Section \ref{sec:algorithm} and assess its performance in Section \ref{sec:evaluation}.

%% file: algorithm.tex
\section{Accuracy aware model partitioning and aggregator assignment algorithm}\label{sec:algorithm}
The proposed algorithm unfolds in two phases.
First, it identifies the set of candidate cut layers that achieve high model accuracy through offline training (Algorithm 1).
Second, it feeds this set as input to a heuristic method that greedily selects the partitioning layers, aggregator layer \& cut layer, and assigns computationally weaker clients to suitable aggregators (Algorithm 2) to minimize the training delay, while satisfying accuracy constraints.
%As the aggregator layer moves deeper into the model, more computational workload is shifted from weaker clients to local aggregators, necessitating an increase in the number of aggregators to maintain low training delay.

\subsection{Algorithm 1: Identification of candidate cut layers}
Before training begins, offline training is conducted for a arbitrary subset of clients $\mathcal{N'}$, a small subset of $\mathcal{N}$, across a range of possible cut layers to assess their impact on the model accuracy.
To facilitate this, an auxiliary network  is attached to the output of every layer, allowing the computation of local-loss gradients and the corresponding accuracy on a per-layer basis.
Specifically, each client trains locally its model up to a candidate cut layer and computes the local-loss gradients, using its local dataset, while the server performs training for the remaining layers.
This offline training procedure is performed for a small number of rounds 
$E'$, after which the resulting model accuracy is measured.
Importantly, the measured accuracy reflects the impact of the chosen cut layer on the overall model accuracy.
Since this procedure is performed over a limited number of rounds, typically smaller than 5, the additional computational overhead is relatively low to the overall training process.

Finally, the server averages these accuracies across all clients to compute an average accuracy score for each candidate cut layer $v$, as depicted in step 8 of Algorithm \ref{alg:cutlayeralgo}.
These scores reflect the ability of each cut layer to achieve high accuracy in local-loss based training.
Since the cut layer selection directly impacts both the computational workload on clients and the achievable accuracy per round, it is important to consider a range of  candidate layers.
Thus, rather than selecting only the single best performing cut layer $v$, the server identifies a set of candidate cut layers, $\mathcal{V^*}$, whose average accuracies are within a tolerance threshold, $thr$, from the maximum observed accuracy.
%The accuracy aware cut layer selection is described also in Algorithm \ref{alg:cutlayeralgo}.

\begin{algorithm}[!tbp]
\caption{Identification of candidate cut layers}\label{alg:cutlayeralgo}
\textbf{Input:} Model: $W$, Clients' datasets: $D_n, \forall n \in \mathcal{N'}$, Number of offline training epochs: $E'$, Accuracy tolerance: $thr$\\
\textbf{Output:} Set of candidate cut layers: $\mathcal{V}^*$
\begin{algorithmic}[1] 
\State Initialize $\text{acc}(v, e) = 0, \forall v \in [2, L{-}1], e \in [1, E']$
\For {$v=2$ to $L-1$}
    \For {$e=1$ to $E'$}
        \For {$n \in \mathcal{N'}$}
            \State Train client $n$ on dataset $D_n$ with cut layer $v$
            \State Compute accuracy $\text{acc}_n(v,e)$ for client $n$
        \EndFor
        \State $\text{acc}(v,e) = \frac{1}{N} \sum_{n \in \mathcal{N}} \text{acc}_n(v,e)$
    \EndFor
    \State $\text{acc}(v, E') = \frac{1}{E'} \sum_{e=1}^{E'} \text{acc}(v,e)$
\EndFor
\State $v^* = \arg\max_{v} \text{acc}(v)$
\State \Return set of candidate cut layers: $\mathcal{V}^* = \{ v \mid \text{acc}(v^*,E') - \text{acc}(v,E') \leq thr \}$
\end{algorithmic}
\end{algorithm}

\subsection{Algorithm 2: Joint selection of partitioning layers and client-to-aggregator assignments}

Given the set of candidate cut layers $\mathcal{V^*}$ computed in Algorithm 1, Algorithm 2 selects a cut layer $v$ within this set that preserves high model accuracy.
The Algorithm 2 then jointly determines the partitioning layers, the aggregator layer $h$ and the cut layer $v$, the subset of clients to operate as local aggregators and the corresponding client-to-aggregator assignments $\mathcal{X}$.
For each cut layer within $\mathcal{V^*}$, it searches the aggregator layer and the fraction of clients operating as local aggregators $\lambda$ such that the selected local aggregators and thus the respective client-to-aggregators assignments minimize the round delay $T_{round}$ (Eq. (\ref{eqn:delay_opt_problem})).
The dominant term in $T_{round}$, as shown in Eq (\ref{eqn:d2}), is determined by the maximum between the delay at the server $T_s$ and the maximum delay over all client-to-aggregator pairs.

When $T_s$ dominates, further balancing the computational workload between clients and aggregators does not further reduce the $T_{round}$.
Otherwise, the bottleneck lies in the client–to-aggregator computation pipeline and in this case, the round delay is minimized when the maximum client-side delay and the maximum aggregator-side delay are approximately balanced (their difference is below a small threshold $\delta$). 
To efficiently identify such an $h$, a binary-search-like procedure is applied over the feasible range $h \in [2, v-1]$.
This is justified by the monotonic trade-off induced by $h$ for fixed $\lambda$: increasing $h$ monotonically increases (weak) client-side delay $T_{clients}$ while decreasing aggregator-side delay $T_{aggr}$.
Consequently, the difference $T_{aggr} - T_{clients}$ changes sign at most once, as shown in step 4 of Algorithm \ref{alg:aggregatorsalgo}, enabling the binary search to converge.

For each value of $\lambda$, the top $\lambda N$ clients, the stronger ones, are nominated as local aggregators.
The remaining $(1-\lambda)N$ clients are assigned to one of these aggregators.
The algorithm computes the expected delay for each client-to-aggregator assignment and selects the assignment that yields the smallest maximum delay across all clients (step 14 in Algorithm 2).
Then, it moves the aggregator layer $h$ deeper or shallower in the model accordingly.

Increasing the fraction of aggregators $\lambda$ introduces a trade-off in $T_{round}$: initially, more aggregators improve workload distribution and reduce client–to-aggregator pipeline delay; however, once the strongest clients have been selected, promoting weaker clients to aggregators yields diminishing returns, as their limited computation throughput increases the delay of the slowest client–to-aggregator pair.
Hence, there exists a threshold $max_{aggr}$ beyond which increasing $\lambda$ cannot reduce, and may even increase the $T_{round}$, and its value is determined by client heterogeneity $\gamma = \max(p_n)/\min(p_n), \forall n \in \mathcal{N}$ and the possible aggregator-side workload.
This threshold also guarantees converge within each iteration.
%Final output
Its output is the partitioning layers $h$, $v$ and the client-to-aggregator assignments $\mathcal{X}$ that reduce the delay as much as possible, while respecting accuracy constraints.
The computation of the client-to-aggregator assignments achieves a worst-case delay that is at most twice the optimal value \cite{lenstra:1990}.

\begin{algorithm}[!tbp]
\caption{Joint selection of partitioning layers and client-to-aggregator assignments}
\label{alg:aggregatorsalgo}
\textbf{Input:} Set of candidate cut layers $\mathcal{V^*}$, client computation throughput $p_n~\forall n \in \mathcal{N}$, transmission rates $r_{i,j}~\forall i,j \in \mathcal{N}$, layer workloads $f_l, \forall l \in \mathcal{V}$, layer bytes $a_l, \forall l \in \mathcal{V}$, heterogeneity ratio $\gamma$, delay threshold $\delta=0.5$, $E$ number of epochs, $B$ batch data size, $D_n, ~\forall n \in \mathcal{N}$ dataset of each client $n$.  \\
\textbf{Output:} Aggregator layer $h$, cut layer $v$, client-to-aggregator assignments $\mathcal{X}$
\begin{algorithmic}[1]
\State Sort clients in descending order of computation throughput.
\For{each cut layer $v \in \mathcal{V^*}$}
\State Initialize $h = 2$, $T_{\text{aggr}} = 10^6$, $T_{\text{clients}} = 0$
\While{$T_{\text{aggr}} - T_{\text{clients}} > \delta$}
    \State Set $maxAggr = \left\lfloor (\gamma - 1) \cdot \frac{\sum_{l=1}^{h} f_l}{\sum_{l=h}^{v} f_l} \right\rfloor$
    \State Set $\lambda_{\max} = \frac{maxAggr}{N}$
    \For{$\lambda = 0.01$ to $\lambda_{\max}$ step $0.01$}
        \State Set top $\lceil\lambda N\rceil$ clients as local aggregators ($\mathcal{K}$)
        \State Initialize $T(k) = 0$,  $\forall k \in \mathcal{K}$.
        \State Initialize assignment matrix $\mathcal{X'}$.
        \For{each client $n \in \mathcal{N} \setminus \mathcal{K}$}
            \For{each aggregator $k \in \mathcal{K}$}
                \State Compute delay $T(k)$ if client $n$ is assigned to aggregator $k$ using eq. (\ref{eqn:d1}), (\ref{eqn:d2}).
            \EndFor
            \State Assign client $n$ to aggregator $k^* = \arg\min_{k \in \mathcal{K}} T(k)$, update $X'$ by assigning layers $(h+1, \ldots, v)$ of client's $n$ model to aggregator $k$.
        \EndFor
        \State Store $h$, $v$, $\mathcal{X'}$ and total delay $T_{round}$, based on eq.(\ref{eqn:dtotal}) using $E$,  $B$, $D_n ~ \forall n \in \mathcal{N}$.
        \State $X = \arg\min_{\mathcal{X'}} T(h,v,\mathcal{X'})$
    \EndFor
    \State $T_{\text{aggr}} = \max_{k \in \mathcal{K}} \left( \frac{\sum_{n=1 } ^ {N} x_{n,k,h} \sum_{l=h+1}^{v} f_l}{p_k} \right)$
    \State $T_{\text{clients}} = \max_ {n \in \mathcal{N}}\left( \frac{\sum_{l=1}^{h} f_l}{p_n} \right)$
    \If{$T_{\text{aggr}} > T_{\text{clients}}$}
        \State  $h = \left\lceil \frac{h + (v-1)}{2} \right\rceil$
    \Else
        \State $h = \left\lceil \frac{h}{2} \right\rceil$
    \EndIf
\EndWhile
\EndFor
\State \Return $(h^*, v^*, \mathcal{X}^*) 
\;=\; \arg\min_{(h,\,v,\,\mathcal{X})} \; T_{\text{round}}$
\end{algorithmic}
\end{algorithm}

\subsection{Algorithm's complexity analysis}
The complexity of Algorithm~\ref{alg:aggregatorsalgo}, which performs a greedy search for the optimal aggregator layer $h$ and client-to-aggregator assignments $\mathcal{X}$, can be broken down as follows.
Initially, the $N$ clients are sorted based on their computation throughputs, which incurs a complexity of $\mathcal{O}(N \log N)$.
The outer loop performs a binary search over the possible aggregator layers $h \in [2 ,v-1 ]$, leading to $\mathcal{O}(\log L)$ iterations.
Within each binary search step, the algorithm evaluates multiple values of the aggregator fraction $\lambda$, incrementally ranging from 0 to a computed upper bound $\lambda_{\max}$.
This upper bound represents the maximum fraction of aggregators for which their inclusion is expected to reduce the overall training delay. The exploration proceeds with a fixed step size $\beta$ (e.g., 0.01).
This results in $\mathcal{O}(1/\beta)$ iterations per $h$.
For each $\lambda$, the algorithm selects the top $\lambda N$ clients as local aggregators and assigns each of the remaining $(1 - \lambda)N$ clients to the aggregator with the minimum estimated delay.
This assignment process requires comparing each (weak) client with every potential aggregator, resulting in a worst case complexity of $\mathcal{O}(N^2)$.
Thus, the total computational complexity of Algorithm~\ref{alg:aggregatorsalgo} is: $\mathcal{O}(\log L  *N^2)$, which is acceptable for moderate scale SFL systems.

\subsection{Discussion on algorithm's robustness to system changes}
Although our proposed solution targets offline optimization decisions (i.e., partitioning layers' selection, client-to-aggregator assignments), in this Section, we discuss how it can be easily adapted in cases where changes in the system occur during the training process.
Such changes may include variations in the transmission rates between clients or the arrival of additional (background) processes (additional computational workload) at the clients that reduce the computational resources available (available computation throughput) for training.
Since the server acts as a centralized orchestrator with a global view of the system, it can recompute and update the system decisions as needed, using the Algorithm \ref{alg:aggregatorsalgo}. 
The selection of the candidate cut layers, determined by Algorithm \ref{alg:cutlayeralgo} is an input parameter to the Algorithm \ref{alg:aggregatorsalgo} and remains constant as it does not rely on system changes, rather than the model architecture.

Due to the nature of the training operations (see Section \ref{sec:system}.B), the selection of the partitioning layers and client-to-aggregator assignments cannot change during the round since the activations \& gradients of the aggregator-side models are stored at the assigned aggregators.
Any system change will be informed to the server during the round, which can run the Algorithm \ref{alg:aggregatorsalgo} in parallel, and broadcast the decisions that will effect the next round.
This shift will last only for a short transition period since re-running Algorithm \ref{alg:aggregatorsalgo} is fast, as demonstrated by the numerical evaluations in Section \ref{sec:evaluation}.D.
As a result, the proposed approach maintains high model accuracy while reducing delay compared to static decision policies, thanks to its ability to efficiently balance computational workloads in response to system changes.
The relevant results are shown and discussed in the Section \ref{sec:evaluation}.E.

%% file: evaluation.tex
\section{Performance evaluation}\label{sec:evaluation}
We denote our solution described in Section \ref{sec:algorithm} as AA HSFL-ll in the following analysis \footnote{The source code will be made available upon paper acceptance.}.
AA HSFL-ll is validated on MNIST \cite{deng:2012}, CIFAR-10 \cite{krizhevsky:2009} and CINIC-10 \cite{darlow:2018} datasets. 
For MNIST, we employ a lightweight neural network consisting of 5 convolutional layers and 3 fully connected layers, following the AlexNet architecture~\cite{alom:2018}, with a total of $W=$ 3,868,170 trainable parameters.
For CIFAR-10, we consider two deeper models.
The first is VGG-11~\cite{iglovikov:2018}, which comprises 8 convolutional layers and 3 fully connected layers, totaling 9,231,114 parameters.
The second one is a heavier model, the VGG-19~\cite{simonyan:2014}, consisting of 16 convolutional layers and 3 fully connected layers, with $W=20,024,394$.
For CINIC-10, a more difficult dataset, we employ ResNet-101 a large residual Residual Network that comprises an initial convolution and 33 bottleneck residual blocks (101 convolutional layers) with $W=44,500,000$ \cite{he:2016}.
In our analysis, we treat each residual block as one layer.
The batch data size is set to 32 and clients perform $E=3$ local epochs per training round.

The training data are distributed i.i.d. across $N=100$ clients, with 30\% are designated as strong clients and the remaining as weak clients.
For experiments involving VGG-19 and ResNet-101, we reduce the number of clients to $N=30$, while maintaining the same proportion of strong clients.
%We expect the qualitative results to persist also for non-i.i.d data across clients\cite{zhao:2018}.
Although non-i.i.d. data distributions may affect convergence behavior \cite{zhao:2018}, the proposed approach, is agnostic to data distribution, as it depends only on computation and communication processes.
The strong clients, such as commercial mobile devices (e.g. Samsung cellphones with Exynos processor) have higher computation throughput, with a total computation throughput of 17.6 GHz (8 cores at 2.2 GHz each) \cite{samsung}.
In contrast, weak clients representative of resource-constrained IoT devices like Raspberry Pis (RPis), e.g. RPi 2, offer a total of 2.4 GHz (2 cores at 1.2 GHz each) \cite{rpi}.
The server's computation throughput $p_s$ equals with 100 Ghz (40 cores of 2.5 Ghz each).
The transmission rate of links $r_{i,j}, \forall i,j \in N$ is uniformly sampled in $[20,25]$ Mbps~\cite{banerjee:2024},~\cite{alliance:2015}.
%Also, for the sake of validity, the computational cost parameters (FLOPs per layer) referred in are measured
%TODO: SAY SOMETHING ABOUT THE client-to-aggregator ASSIGNMENTS, GIVE SOME NUMBERS, DISCUSS IT.
%Maybe add a table?
%also vgg 19 has a bigger win in overhead since it is a bery larger model than the others\
%Mention fewer clients for vgg-19
\subsection{Accuracy, delay and overhead evaluation}
In this section, we evaluate the performance of the proposed AA HSFL-ll approach.
The first set of experiments compares AA HSFL-ll with representative SFL schemes in terms of model accuracy, training delay, and communication overhead:
\begin{itemize}
    \item SFL: The model $W$ is split at the cut layer $v$ into the client-side $W^c$ and server-side $W^s$ models.
    Both of them are trained in a sequential manner, the clients wait for the gradients from the server to perform BP.
The source code provided by the authors is reused~\cite{thapa:2022}.
 \item Multihop SFL: The model $W$ is split at multiple cut layers, enabling a hierarchical computation structure where each client trains different portions of its own model or assists weaker clients in a pipeline based manner~\cite{Zheng:2025}.
 \item LocSFL: The client-side $W^c$ and server-side $W^s$ models are trained in parallel by employing a local-loss at the cut layer $v$, allowing clients to perform BP without waiting for gradients from the server~\cite{han:2021}.
\item DTFL: Similarly, $W^c$ and $W^s$ are trained in parallel using local-loss at the cut layer and each client is assigned a personalized cut layer based on its computation throughput~\cite{mohammadabadi:2024}.
\end{itemize}
Notably, the aforementioned schemes do not consider the impact of cut layer selection to the model accuracy.
\begin{figure}
\begin{tabular}{c c}
\hspace{-0.2cm}\includegraphics[angle=0,trim={0cm 0cm 0cm 0cm},clip,scale=0.31]{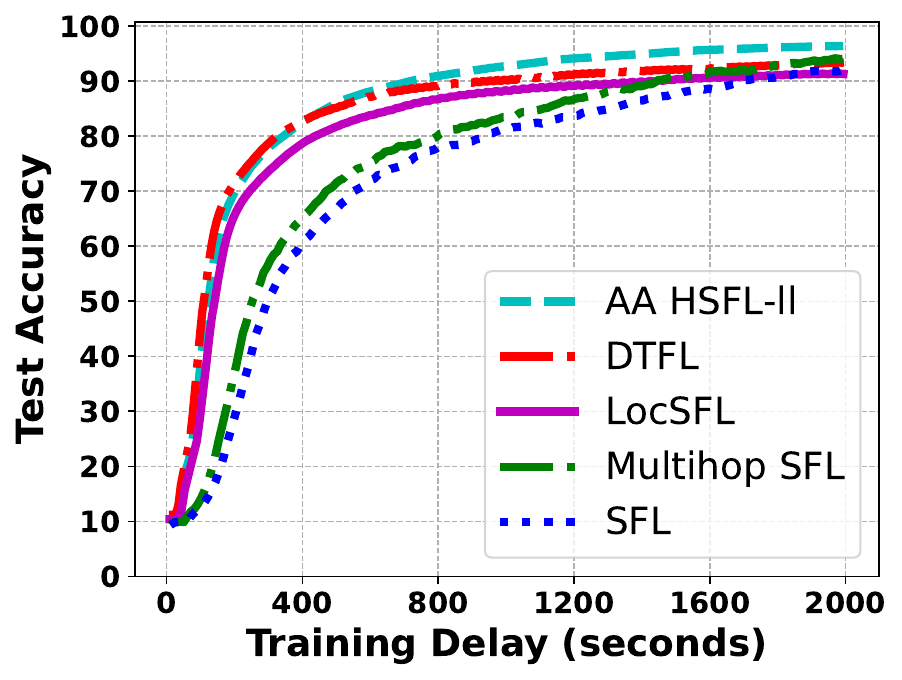} & \hspace{-0.3cm}\includegraphics[angle=0,trim={2cm 0cm 0cm 0cm},clip,scale=0.31]{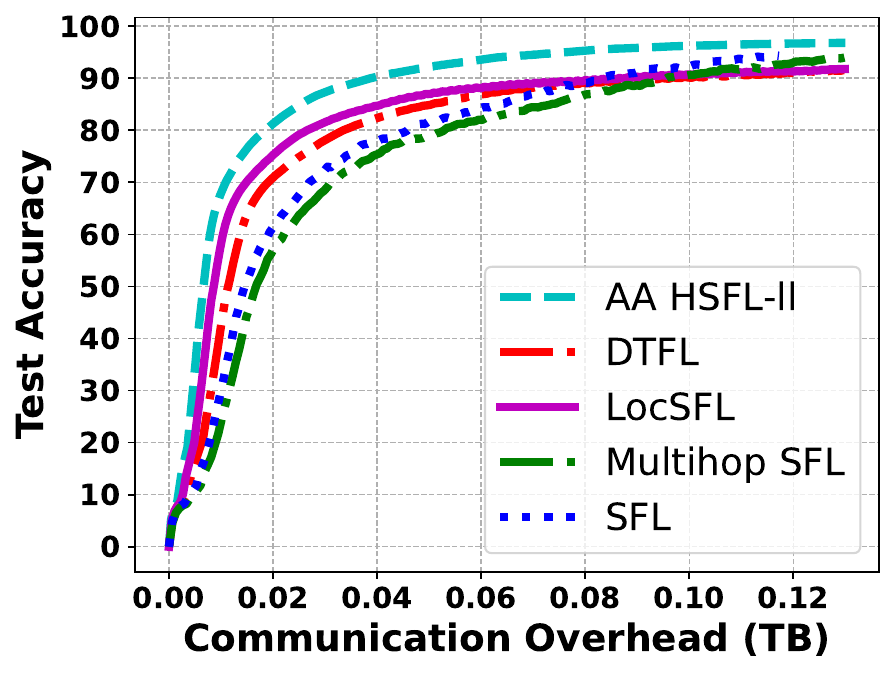} \\
\hspace{-1.1cm} \footnotesize (a) Test accuracy vs delay & \hspace{-1.0cm} \footnotesize (b) Test accuracy vs overhead  \\
\end{tabular}
\caption{Test accuracy versus (a) training delay and (b) communication overhead achieved by AA HSFL-ll and baseline schemes during the training of the AlexNet model on the MNIST dataset.}
\label{fig:alexnet_exp1}
\vspace{-0.3cm}
\end{figure}

\begin{figure}
\begin{tabular}{c c}
\hspace{-0.2cm}\includegraphics[angle=0,trim={0cm 0cm 0cm 0cm},clip,scale=0.31]{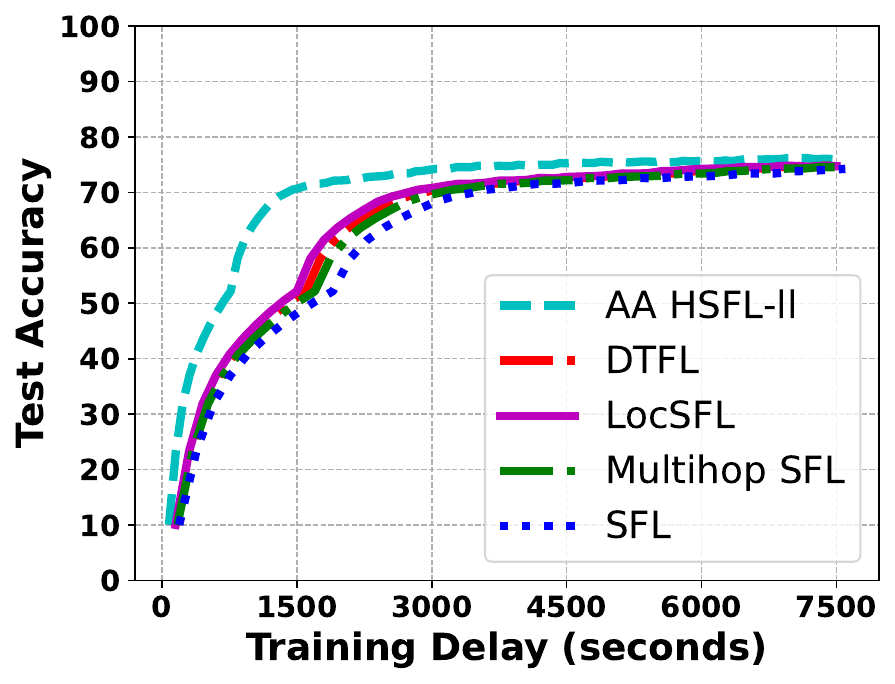} & \hspace{-0.3cm}\includegraphics[angle=0,trim={2cm 0cm 0cm 0cm},clip,scale=0.31]{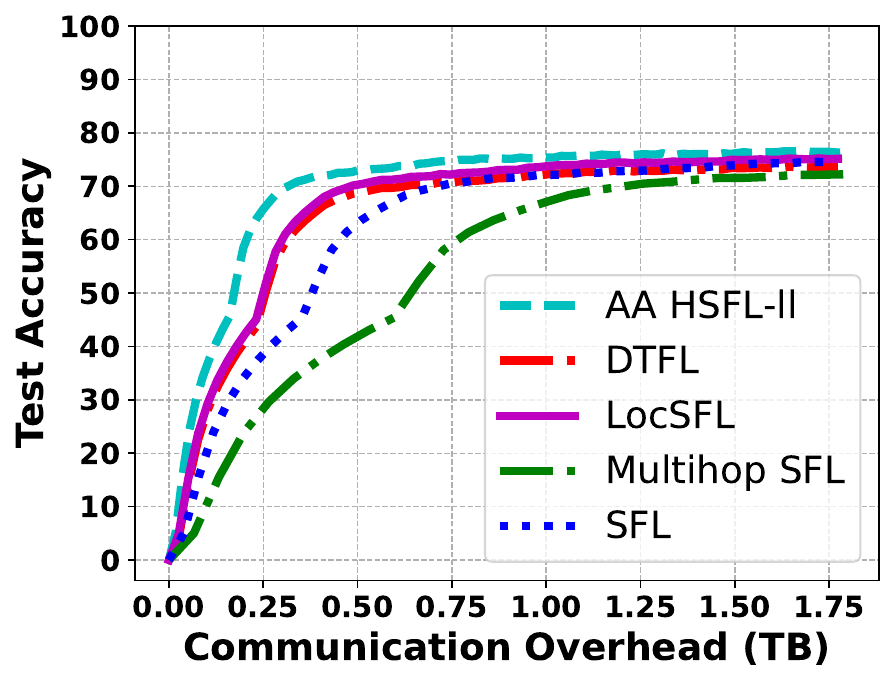} \\
\hspace{-1.1cm} \footnotesize (a) Test accuracy vs delay & \hspace{-1.0cm} \footnotesize (b) Test accuracy vs overhead  \\
\end{tabular}
\caption{Test accuracy versus (a) training delay and (b) communication overhead achieved by AA HSFL-ll and baseline schemes during the training of the VGG-11 model on the CIFAR-10 dataset.}
\label{fig:vgg11_exp1}
\vspace{-0.3cm}
\end{figure}
%insert the Ks into the plots of the figures
\begin{figure}
\begin{tabular}{c c}
\hspace{-0.2cm}\includegraphics[angle=0,trim={0cm 0cm 0cm 0cm},clip,scale=0.31]{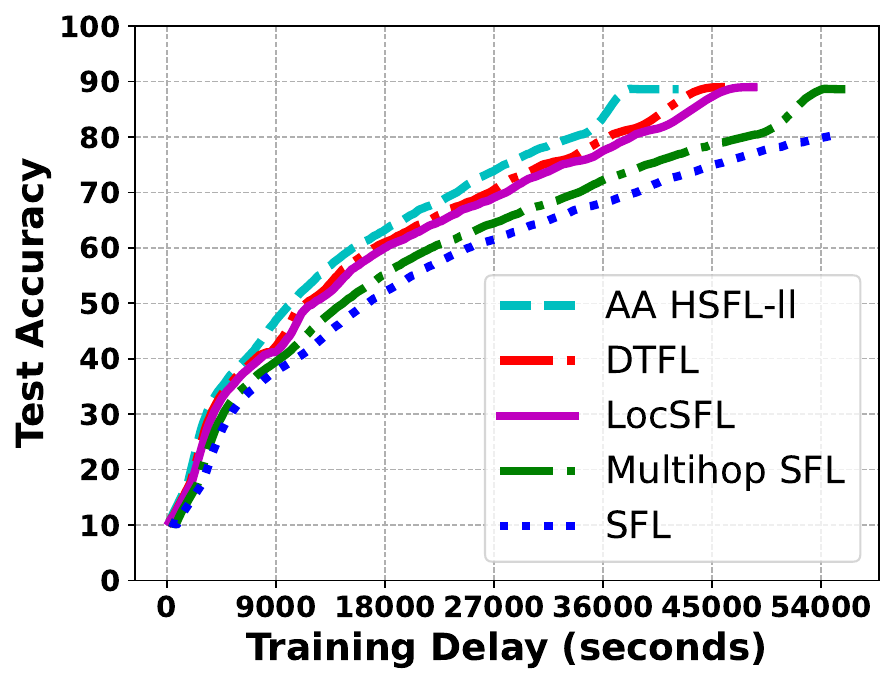} & \hspace{-0.3cm}\includegraphics[angle=0,trim={2cm 0cm 0cm 0cm},clip,scale=0.31]{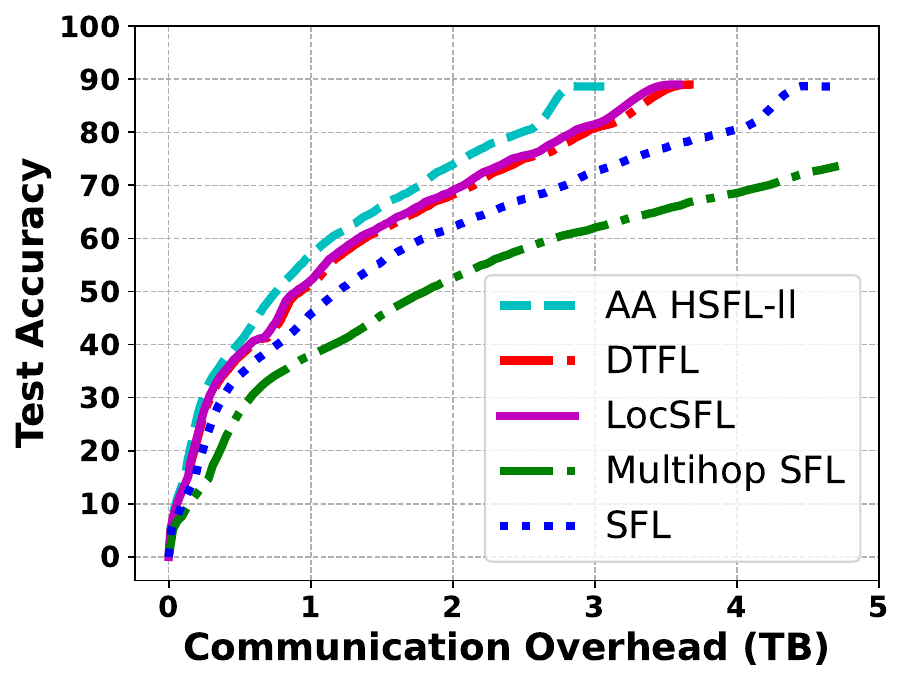} \\
\hspace{-1.1cm} \footnotesize (a) Test accuracy vs delay & \hspace{-1.0cm} \footnotesize (b) Test accuracy vs overhead  \\
\end{tabular}
\caption{Test accuracy versus (a) training delay and (b) communication overhead achieved by AA HSFL-ll and baseline schemes during the training of the VGG-19 model on the CIFAR-10 dataset.}
\label{fig:vgg19_exp1}
\vspace{-0.3cm}
\end{figure}
%Make the table in terms of gains!
\begin{figure}
\begin{tabular}{c c}
\hspace{-0.5cm}\includegraphics[angle=0,trim={0cm 0cm 0cm 0cm},clip,scale=0.31]{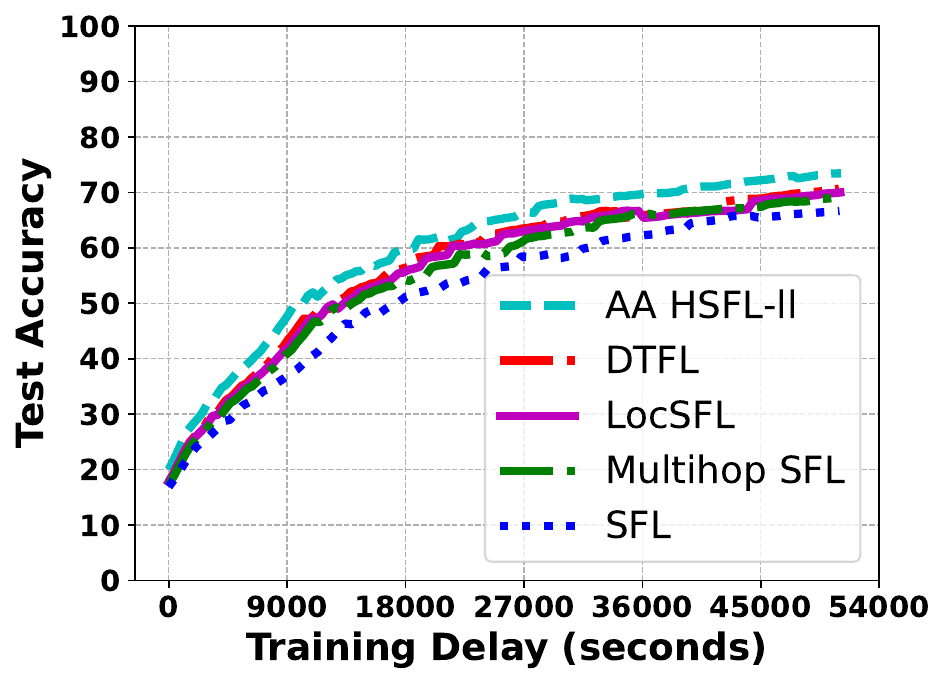} & \hspace{-0.3cm}\includegraphics[angle=0,trim={2cm 0cm 0cm 0cm},clip,scale=0.31]{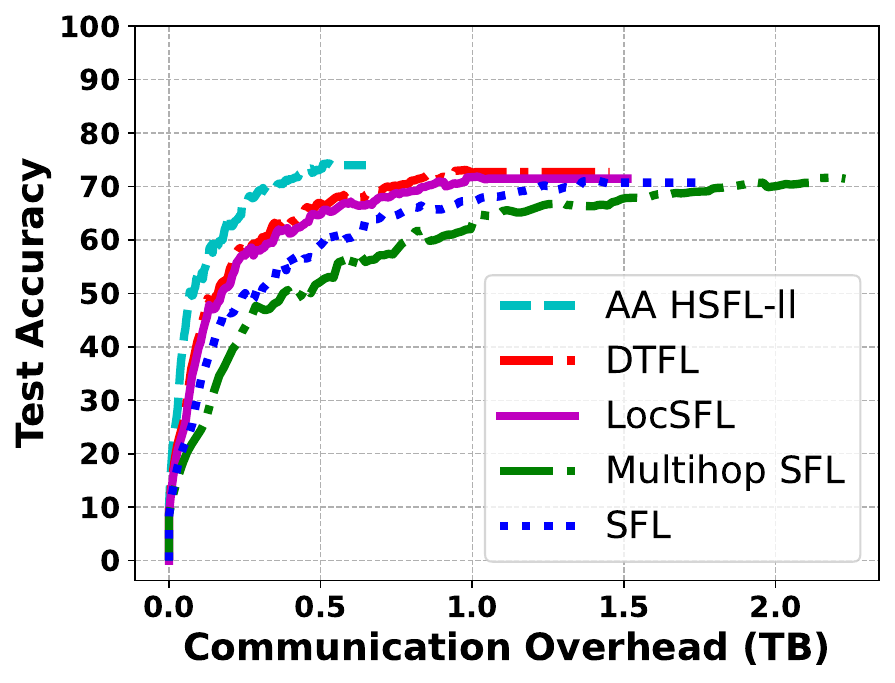} \\
\hspace{-1.1cm} \footnotesize (a) Test accuracy vs delay & \hspace{-1.0cm}
\footnotesize (b) Test accuracy vs overhead  \\
\end{tabular}
\caption{Test accuracy versus (a) training delay and (b) communication overhead achieved by AA HSFL-ll and baseline schemes during the training of the ResNet-101 model on the CINIC-10 dataset.}
\label{fig:resnet_exp1}
\vspace{-0.3cm}
\end{figure}
\begin{table}[tb]
\caption{Relative gains (\%) achieved by AA-HSFL-ll in terms of training delay and communication overhead for various accuracy targets compared to the best performing baseline scheme.}
\centering
\footnotesize
\begin{tabular}{|c|c|c|c|}
\hline
Model & Acc. Target(\%) & Delay Gain (\%) & Overhead Gain (\%) \\
\hline
AlexNet & 85 & 7.25 & 57.14 \\
AlexNet & 90 & 22.00 & 55.56 \\
AlexNet & 94 & 20.00 & 50.00 \\
\hline
VGG-11 & 65 & 35.00 & 37.50 \\
VGG-11 & 70 & 28.41 & 30.00 \\
VGG-11 & 75 & 11.34 & 34.62 \\
\hline
VGG-19 & 80 & 5.48 & 21.05 \\
VGG-19 & 85 & 11.00 & 19.40 \\
VGG-19 & 90 & 8.22 & 22.08 \\
\hline
ResNet-101 & 65 & 28.22 & 45.45 \\
ResNet-101 & 70 & 19.42 & 50.00 \\
ResNet-101 & 73 & 8.25 & 35.29 \\
\hline
\end{tabular}
\label{tab:target_accuracyTable}
\end{table}
%For the AlexNet, VGG-11, VGG-19 and ResNet-101 models, convolutional layers 5, 8, 15, and 24 respectively, are selected as the cut layers, as these choices yield the highest per-round accuracy when applying Algorithm 1 (Section \ref{sec:algorithm}).
%To ensure a fair comparison, the same optimal cut layer is used for both the LocSFL and SFL schemes.
%Figures \ref{fig:alexnet_exp1}, \ref{fig:vgg11_exp1}, and \ref{fig:vgg19_exp1} illustrate the achieved accuracy of all schemes as a function of training delay and communication overhead for the three models.
In Figs. \ref{fig:alexnet_exp1}, \ref{fig:vgg11_exp1}, \ref{fig:vgg19_exp1} and \ref{fig:resnet_exp1} we plot the achieved accuracy of each scheme as a function of the training delay and the communication overhead for the four models.
Table \ref{tab:target_accuracyTable} shows the relative gains in training delay and communication overhead for different accuracy targets across the four models.
As shown in Fig. \ref{fig:alexnet_exp1}a, AA HSFL-ll attains the highest accuracy during AlexNet training within 2000 seconds.

%The results demonstrate that AA HSFL-ll consistently outperforms the best performing alternative scheme across all accuracy targets.
To reach 94\% accuracy on AlexNet, AA HSFL-ll requires 1600 seconds, compared to 2000 seconds for Multihop SFL, corresponding to a 20\% reduction in delay (see Table \ref{tab:target_accuracyTable}).
Within the same 1600 seconds training duration, AA HSFL-ll achieves 94\% accuracy, while the best competing scheme, DTFL, attains only 91\%, demonstrating a 3\% accuracy improvement for our approach.
Also as shown in Fig. \ref{fig:alexnet_exp1}b, for the same accuracy target, AA HSFL-ll reduces the overhead by approximately 50\% compared to Multihop SFL (0.06 versus 0.12 TB). 
Similarly, for VGG-11 at 70\% accuracy target, AA HSFL-ll reduces the delay by 28\% compared to LocSFL (1580 versus 2210 seconds), as well as the overhead by 30\% (0.35 versus 0.5 TB).

For the deeper VGG-19 model, the advantages of AA HSFL-ll become even more pronounced. At 85\% accuracy target, AA HSFL-ll achieves a 11\% reduction in delay compared to DTFL (37200 versus 41800 seconds), while simultaneously reducing overhead by approximately 20\% (see Fig.~\ref{fig:vgg19_exp1}).
This confirms that aggressively offloading computations to the server, as in DTFL, is not always efficient for deep models with high computation and communication demands.

For the ResNet-101 model, similar trends are observed, as illustrated in Fig. \ref{fig:resnet_exp1}.
AA HSFL-ll consistently reaches target accuracies with lower delay and overhead than the rest schemes.
For example, at 70\% accuracy target, AA HSFL-ll reduces delay by approximately 20\% compared to DTFL (36,100 seconds versus 44,800 seconds).
As shown in Fig. \ref{fig:resnet_exp1}a and Fig. \ref{fig:resnet_exp1}b, AA HSFL-ll exhibits faster accuracy growth with respect to both delay and overhead, highlighting its effectiveness handling deep residual model architectures.

The results demonstrate that AA HSFL-ll consistently outperforms the best performing alternative scheme across all accuracy targets.
The superior performance of AA HSFL-ll can be attributed to its adaptive partitioning and client-to-aggregator assignment strategy.
For smaller values of $\lambda$, fewer local aggregators, more models are aggregated within the same local aggregators, which improves per round accuracy, while avoiding additional overhead.
An additional communication cost is incurred due to the exchange of activations and gradients at the additional layer $h$.
However, this overhead remains relatively small compared to the communication cost associated with transmitting entire sub-models.
Moreover, the adaptive selection of the aggregator layer $h$ lightens the computational workload from weaker clients, thereby reducing overall delay.

In contrast, both SFL and Multihop SFL incur larger delays because clients remain idle while waiting for gradients from the server (due to the backward-locking effect).
Although DTFL mitigates the backward-locking effect using local-loss gradients, its use of shallow cut layers (close to the first layer) for weak clients shifts substantial computation workload to the server, thereby severely increasing delay.
This observation highlights a key advantage of AA HSFL-ll: rather than offloading computation workloads entirely to the server, it balances this workload between clients and local aggregators in a delay-aware manner.

Finally, Table \ref{tab:decisionTable} summarizes the decisions made by the proposed algorithm for each model, including the selected cut and aggregator layers, number of aggregators, and average clients assigned per aggregator.
As model depth increases, the algorithm places both layers deeper in the network and increases the aggregator ratio $\lambda$, resulting in fewer clients per aggregator.
For instance, from VGG-11 to ResNet-101, the cut layer shifts from $v=8$ to $v=24$ and $\lambda$ increases from 0.20 to 0.30, reducing the average number of clients assigned per aggregator from 4 to 2.25.
%This behavior reflects the algorithm’s ability to balance computational workload and communication overhead, and helps explain the consistent performance gains of AA HSFL-ll across models of increasing complexity.
This adaptive behavior explains the consistent gains of AA HSFL-ll across models of increasing complexity.
\begin{table}[tb]
\caption{Selected aggregator layer ($h$), cut layer ($v$) fraction of clients operating as local aggregators ($\lambda$), and average clients assigned per aggregator during the training of the four models using the proposed algorithm.}
\centering
\footnotesize
\begin{tabular}{|c|c| c| c|c|}
\hline
  Model &  $h$ & $v$ & $\lambda$ & Avg. clients \\
\hline
    AlexNet  & $3$ & $5$  &   $0.2$ &   5.5\\
    \hline
    VGG-11 &  $3$ &$8$     &   $0.2$&  4 \\
    \hline
    VGG-19 &  $4$ & $15$     &   $0.25$ & 3 \\
    \hline
    ResNet-101 &  $8$ &$24$     &   $0.3$ & 2.25 \\
    \hline
\end{tabular}
\label{tab:decisionTable}
\end{table}
\subsection{Effect of fraction of clients operating as aggregators $\lambda$}
\begin{figure}
\begin{tabular}{c c}
\hspace{-0.3cm}\includegraphics[angle=0,trim={0cm 0cm 0cm 0cm},clip,scale=0.31]{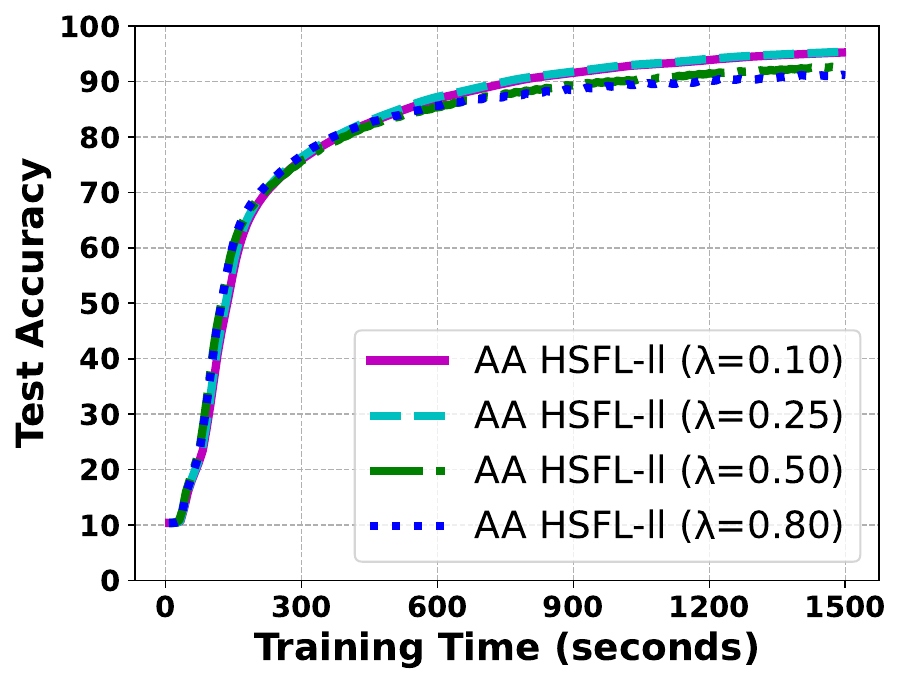} & \hspace{-0.3cm}\includegraphics[angle=0,trim={2cm 0cm 0cm 0cm},clip,scale=0.31]{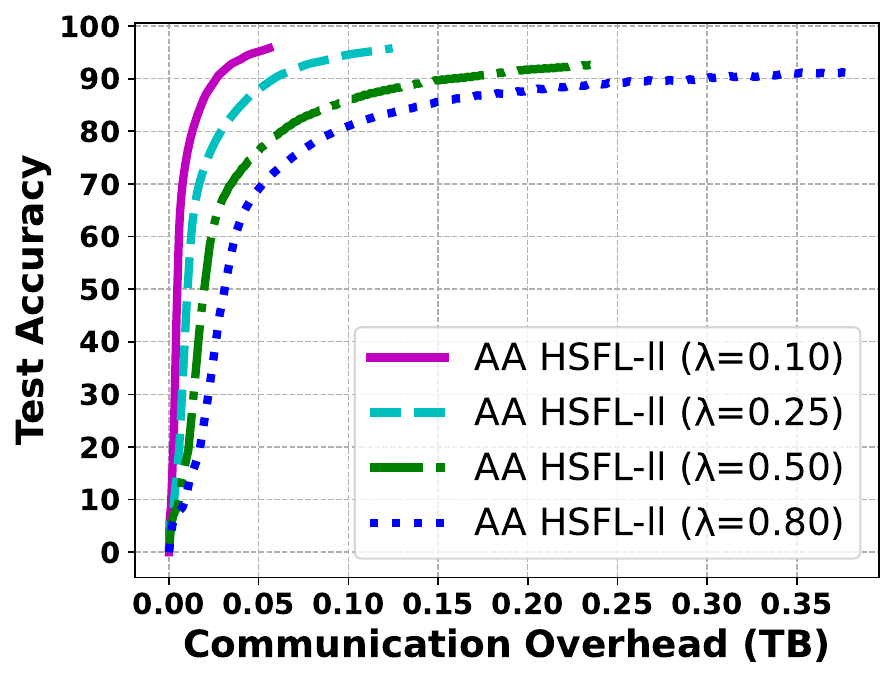} \\
\hspace{-1.1cm} \footnotesize (a) Test accuracy vs delay & \hspace{-0.0cm} \footnotesize (b) Test accuracy vs overhead \\
\end{tabular}
\caption{Test accuracy obtained by AA HSFL-ll during training the AlexNet model for various numbers of local aggregators $\lambda N$.}
\label{fig:alexnet_exp2}
\vspace{-0.3cm}
\end{figure}

\begin{figure}
\begin{tabular}{c c}
\hspace{-0.3cm}\includegraphics[angle=0,trim={0cm 0cm 0cm 0cm},clip,scale=0.31]{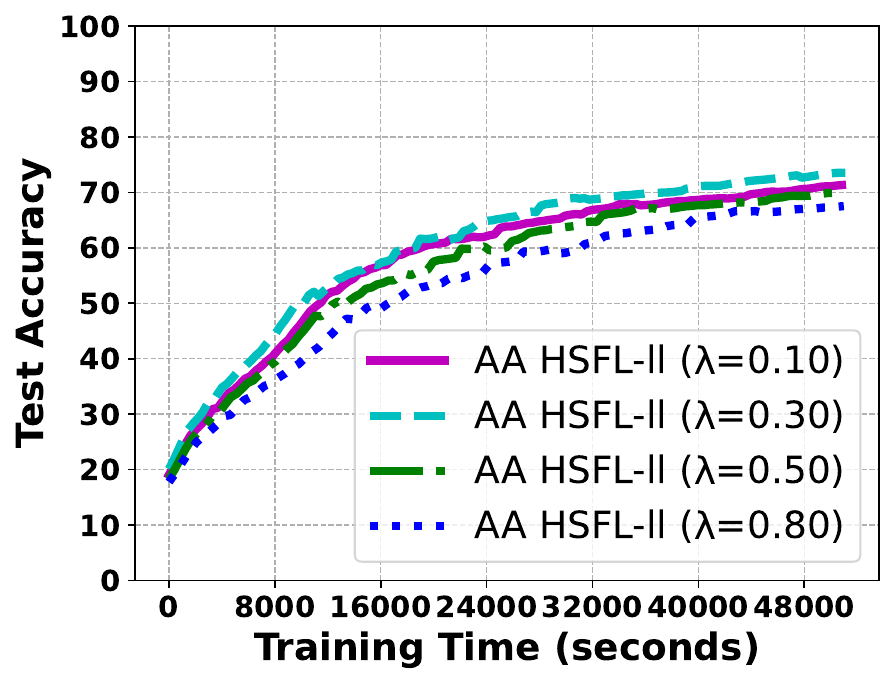} & \hspace{-0.3cm}\includegraphics[angle=0,trim={2cm 0cm 0cm 0cm},clip,scale=0.31]{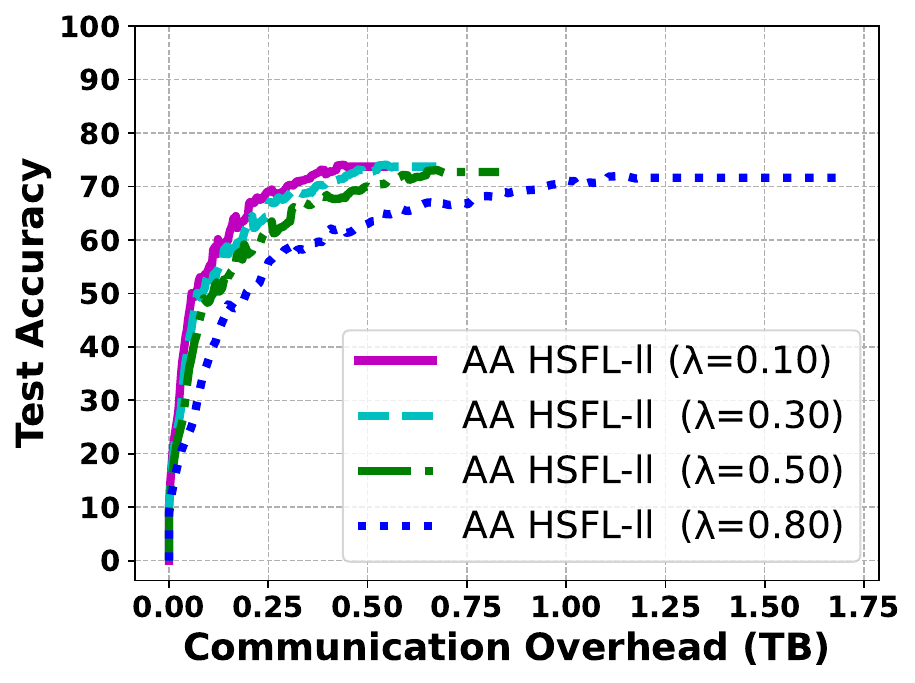} \\
\hspace{-1.1cm} \footnotesize (a) Test accuracy vs delay & \hspace{-0.0cm} \footnotesize (b) Test accuracy vs overhead \\
\end{tabular}
\caption{Test accuracy versus communication overhead obtained by AA HSFL-ll during training the ResNet-101 model for various numbers of local aggregators $\lambda N$.}
\label{fig:resnet_exp2}
\vspace{-0.3cm}
\end{figure}

Previous experiments employed Algorithm 2 to determine the number of local aggregators.
Here, we study the impact of different values of $\lambda$, portion of clients nominated as local aggregators, on AA HSFL-ll during the training of AlexNet and ResNet-101 models.
As shown in Fig. \ref{fig:alexnet_exp2}a, increasing $\lambda$ from 0.10 to 0.25 yields only marginal accuracy gains (below 1\%) for AlexNet during the same time of training, whereas for ResNet-101, increasing $\lambda$ from 0.10 to 0.30 improves accuracy by up to 3\% (Fig. \ref{fig:resnet_exp2}a).
In terms of overhead, AlexNet and ResNet-101 incur higher overhead for larger values of $\lambda$ when achieving the same accuracy (see Fig. \ref{fig:alexnet_exp2}b \& Fig.\ref{fig:resnet_exp2}b).
As summarized in Table \ref{tab:layersTable}, the aggregator layer $h$ adapts with $\lambda$, shifting from $h=3$ to $h=2$ for AlexNet, while the cut layer remains the same $v=5$.
For ResNet-101, the aggregator layer moves from $h=8$ to $h=12$ and the cut layer from $v=24$ to $v=26$, indicating a redistribution of computation workload toward local aggregators as the number of aggregators is increased (bigger $\lambda$).
Notably, Algorithm 2 selects the partitioning layers that achieve the best trade-off between accuracy and training delay, while maintaining the communication overhead close to its minimum feasible level.

However, when $\lambda$ exceeds 0.50, no further reductions in delay are observed, and model accuracy may degrade by up to 4\%.
This degradation stems from the diminishing effect of local aggregations.
For instance, at $\lambda = 0.10$, each local aggregator may aggregate from up to 10 aggregator-side models, while at $\lambda = 0.80$, some aggregators may aggregate from only 2–3 models or even only their own model, leading to ineffective local aggregation.
When the parameter $\lambda$ is set to $\lambda=1$, AA HSFL-ll reduces to the LocSFL scheme, where each client aggregates only its own model (practically no aggregation occurs at the local aggregator).
Simultaneously, overhead increases with higher values of $\lambda$, as more aggregator-side models are transmitted.
Specifically, as shown in Fig.~\ref{fig:alexnet_exp2}b, AA HSFL-ll with $\lambda = 0.10$ requires only 0.03 TB of communication overhead to achieve nearly 90\% accuracy, whereas 0.15 TB and 0.2 TB are required when $\lambda \in \{0.50, 0.80\}$, respectively.
We observe a similar trend for the ResNet-101 training in Fig. \ref{fig:resnet_exp2}b. 

\begin{comment}
\begin{figure}
\begin{tabular}{c c}
\hspace{-0.3cm}\includegraphics[angle=0,trim={0cm 0cm 0cm 0cm},clip,scale=0.31]{figures/aggregators_delay_vgg11.pdf} & \hspace{-0.3cm}\includegraphics[angle=0,trim={2cm 0cm 0cm 0cm},clip,scale=0.31]{figures/aggregators_overhead_vgg11.pdf} \\
\hspace{-1.1cm} \footnotesize (a) Test accuracy vs delay & \hspace{-0.0cm} \footnotesize (b) Test accuracy vs overhead \\
\end{tabular}
\caption{\textbf{ TO BE REPLACED}Test accuracy versus communication overhead obtained by AA HSFL-ll during training the ResNet101 model for various numbers of local aggregators $\lambda N$.}
\label{fig:vgg11_exp2}
\vspace{-0.3cm}
\end{figure}
\end{comment}

\begin{table}[tb]
\caption{Aggregator layer and cut layer ($h$, $v$) selection during the training of the AlexNet and ResNet-101 models for various fractions of clients operating as local aggregators $\lambda$.}
\centering
\footnotesize
\begin{tabular}{|c|c|c|c| c| c|}
\hline
 $\lambda$ & 0.10 &  0.25& 0.30 & 0.50 & 0.80\\
\hline
    AlexNet  & (3, 5) & (2, 5) & - & (2, 5)  & (2, 5)  \\
     \hline
    ResNet-101 & (9, 24) & - & (8, 24) & (9, 26) & (12, 26)    \\
    \hline
\end{tabular}
\label{tab:layersTable}
\end{table}
\subsection{Effect of client heterogeneity $\gamma$}
We next investigate the impact of client heterogeneity, quantified by the heterogeneity degree $\gamma$, on the performance of AA HSFL-ll.
To this end, we repeat the experiments of the first group while focusing exclusively on AA HSFL-ll for the VGG-11 and ResNet-101 models, and consider three heterogeneity scenarios: low heterogeneity ($\gamma=2$), medium heterogeneity ($\gamma=7.5$), as in the first group of experiments in Section~\ref{sec:evaluation}.A, and high heterogeneity ($\gamma=15$).
As reported in Table \ref{tab:heterogeneityTable}, the achieved accuracy remains stable across different heterogeneity ratios for both models.
Particularly, VGG-11 exhibits only marginal accuracy variations, while ResNet-101 shows similar performance across all values of $\gamma$, indicating that AA HSFL-ll is robust to varying client heterogeneity.

As client heterogeneity increases, the algorithm adaptively changes its output.
Specifically, Table \ref{tab:heterogeneity1Table} shows that higher values of $\gamma$ lead to an increase in the number of local aggregators ($\lambda N$), reflecting a shift toward more distributed local aggregation.
At the same time, the aggregator layer $h$ moves deeper into the network for VGG-11 (from $h=3$ to $h=5$), while the cut layer remains the same $v=8$.
For ResNet-101, the aggregator-layer shifts toward earlier layers (from $h=11$ to $h=6$) and the cut layer from $v=26$ to $v=24$.
This adaptive behavior enables effective workload redistribution from weaker clients to local aggregators as heterogeneity grows, highlighting the importance of AA HSFL-ll in practical deployments involving highly heterogeneous and resource-constrained devices, such as Raspberry PIs.
%\begin{comment}
% \begin{figure}
% \begin{tabular}{c c}
% \hspace{-0.5cm}\includegraphics[angle=0,trim={0cm 0cm 0cm 0cm},clip,scale=0.31]{figures/aggregators_delay_mnist.pdf} & \hspace{-0.3cm}\includegraphics[angle=0,trim={2cm 0cm 0cm 0cm},clip,scale=0.31]{figures/aggregators_overhead_mnist.pdf} \\
% \hspace{-1.1cm} \footnotesize (a) AlexNet & \hspace{-0.0cm} \footnotesize (b) VGG-11 placeholder\\
% \end{tabular}
% \caption{Accuracy obtained by AA HSFL-ll during training the AlexNet and VGG-11 models for various client heterogeneity degrees $\gamma$.}
% \label{fig:delay_exp3}
% \vspace{-0.3cm}
% \end{figure}
% \begin{figure}
% \begin{tabular}{c c}
% \hspace{-0.5cm}\includegraphics[angle=0,trim={0cm 0cm 0cm 0cm},clip,scale=0.31]{figures/aggregators_delay_mnist.pdf} & \hspace{-0.3cm}\includegraphics[angle=0,trim={2cm 0cm 0cm 0cm},clip,scale=0.31]{figures/aggregators_overhead_mnist.pdf} \\
% \hspace{-1.1cm} \footnotesize (a) AlexNet & \hspace{-0.0cm} \footnotesize (b) VGG-11 placeholder \\
% \end{tabular}
% \caption{Communication overhead obtained by AA HSFL-ll during training the AlexNet and VGG-11 models for various client heterogeneity degrees $\gamma$.}
% \label{fig:overhead_exp3}
% \vspace{-0.3cm}
% \end{figure}
%\end{comment}
\begin{table}[tb]
\caption{Accuracy achieved by our approach for various client heterogeneity degrees $\gamma$ during the training of the VGG-11 and ResNet-101 models.}
\centering
\footnotesize
\begin{tabular}{|c|c|c|}
\hline
   Model & $\gamma$ & Accuracy (\% )  \\
\hline
    VGG-11 & $2$  & 76    \\
    VGG-11 & $7.5$ & 76.8  \\
    VGG-11 & $15$ &   75.2  \\
    \hline
    ResNet-101  & $2$ & 73     \\
    ResNet-101  & $7.5$ & 73.2   \\
    ResNet-101  & $15$ & 73    \\
    \hline
\end{tabular}
\label{tab:heterogeneityTable}
\end{table}
\begin{table}[tb]
\caption{Aggregator layer $h$ and fractions of clients operating as local aggregators $\lambda$ selection during the training of the VGG-11 and ResNet-101 models for various client heterogeneity degrees $\gamma$.}
\centering
\footnotesize
\begin{tabular}{|c|c|c|c|c|}
\hline
  Model &  $\gamma$ &  $h$ & $v$&  $\lambda$  \\
\hline
    VGG-11 & $2$  & $3$   & $8$   &    $0.10$ \\
    VGG-11 & $7.5$ & $4$   &  $8$   & $0.20$ \\
    VGG-11 & $15$ & $5$   &  $8$   &  $0.30$\\
    \hline
    ResNet-101  & $2$ &   $11$  & $26$   & $0.15$ \\
    ResNet-101  & $7.5$ & $8$ & $24$   &  $0.30$ \\
    ResNet-101  & $15$ & $6$   &  $24$   & $0.30$ \\
    \hline
\end{tabular}
\label{tab:heterogeneity1Table}
\end{table}

\subsection{Approximation to the optimal solution}
In Section \ref{sec:algorithm}, we demonstrated that the problem is NP-hard and proposed a heuristic algorithm as a tractable solution.
A natural question arises: how close is our heuristic to the optimal solution?
To address this, we evaluate the suboptimality gap and the execution time speedup achieved by our algorithm \ref{alg:aggregatorsalgo} compared to exhaustive search for various number of clients $N$ through offline simulations.

As shown in Table~\ref{tab:speedupTable}, the proposed algorithm achieves suboptimality below 12\% across all configurations while significantly reducing execution time.
For medium problem sizes ($N=30, 60$), the heuristic consistently attains optimal solutions for both VGG-19 and ResNet-101, while already providing speedups between 10$\times$ and 24$\times$ over exhaustive search.
As the number of clients increases to $N=100$, exhaustive search becomes prohibitively expensive, whereas the proposed algorithm continues to scale efficiently, achieving speedups of up to 36$\times$ and 40$\times$, respectively.
Although a moderate increase in suboptimality is observed in these large-scale settings, the resulting solutions remain close to optimal, highlighting a favorable trade-off between solution optimality and computational efficiency in medium/large-scale systems.
\begin{table}[tb]
\caption{Sub-optimality and speedup achieved by
our algorithm when compared to the exhaustive search for various number of clients $N$ during the training of VGG-19 and ResNet-101 models.}
\centering
\footnotesize
\begin{tabular}{|c|c|c| c|}
\hline
  Model &  $N$ & sub-optimality (\%) & speedup ($x$ Times)\\
\hline
    VGG-19 & 30 & 0 & 10.5   \\
    VGG-19 & 60 & 0 & 24   \\
    VGG-19 & 100 & 10.5 & 36   \\
    \hline
    ResNet-101  & 30 & 0 & 10.5   \\
    ResNet-101  & 60 & 0& 22     \\
    ResNet-101  & 100 & 12 & 40     \\
    \hline
\end{tabular}
\label{tab:speedupTable}
\end{table}

\subsection{Results on robustness to  system changes}
Following the discussions in Section~\ref{sec:algorithm}.D concerning the proposed solution’s robustness and adaptation to system changes, we explore the impact of system variations on the delay.
%Since robustness is not the main focus of this work, we present preliminary results that demonstrate the ability of the proposed method to adapt to changing conditions and reduce the inevitable performance degradation as much as possible.
We first focus on the rows of Table~\ref{tab:changeComparison} which examine scenarios that additional processing tasks arrive at the client devices.
In these experiments, the computational resources that are available for training are reduced by up to 30\% due to competing workloads (background tasks and training).
As expected, such changes lead to increased training delay compared to the original system configuration.
However, the results show that dynamically recomputing the solution significantly limits this increase. For example, when 30\% additional tasks are introduced, the delay increase for VGG-19 is reduced from 12\% under the fixed solution to 5\% with the proposed adaptive solution (New sol. in Table \ref{tab:changeComparison}).
Similar trends are observed for ResNet-101, where the delay increase is reduced from 24\% to 11\%.

We next consider the rows of Table~\ref{tab:changeComparison} that examine scenarios with a severely reduced transmission rate (below 4~Mbps).
In this case, communication becomes the dominant delay bottleneck, particularly for models with larger layer representations.
For VGG-19, the fixed solution suffers an 18\% delay increase, while the adaptive solution reduces this to 7\%.
For ResNet-101, which generates smaller tensors per layer, the impact of reduced transmission rate is less pronounced, yet the adaptive solution still consistently outperforms the fixed one.

%Naturally, system changes typically occur gradually over time. For simplicity, the experiments in Table~\ref{tab:changeComparison} assume that multiple changes occur simultaneously, resulting in relatively large delay variations.
%Despite this setting, the results indicate that the dynamic approach can effectively respond to system changes and effectively reduce the resulting delay increase.
\begin{table}[tb]
\caption{Comparison of fixed and proposed solutions under different change types during the training of VGG-19 and ResNet-101 models.}
\centering
\footnotesize
\begin{tabular}{|c|c|c|c|}
\hline
Type of change& Model& Fixed sol. (\%) & New sol. (\%) \\
\hline
Extra 15 \% tasks  & VGG-19 &  +5  &  +2  \\
Extra 30 \%  tasks   & VGG-19  &  +12   &  +5 \\
Reduced TX rate   & VGG-19  &  +18    & +7   \\
\hline
Extra 15 \% tasks  & ResNet-101 &  +14   &  +9  \\
Extra 30 \%  tasks   & ResNet-101  &  +24  &  +11  \\
Reduced TX rate  & ResNet-101  &  +7  &  +3  \\
\hline
\end{tabular}
\label{tab:changeComparison}
\end{table}

%% file: conc.tex
\section{Conclusion}\label{sec:conclusion}
%In this work, we proposed a novel algorithm for distributed, privacy-preserving training of machine learning models under the Split Federated Learning (SFL) paradigm.
%
In this work, we studied the joint optimization of model partitioning and client-to-aggregator assignment in Hierarchical Split Federated Learning (HSFL) schemes, with the goal of minimizing training delay while accounting for model accuracy.
We formulated the problem as a joint optimization over the partitioning layers and client-to-aggregator assignments, proved its NP-hardness, and proposed the first accuracy-aware heuristic algorithm that efficiently improves model accuracy while reducing training delay and communication overhead.
In particular, our algorithm significantly reduces training delay by 20\% and communication overhead by 50\%, while improving model accuracy by 3\% compared to state-of-the-art SFL and HSFL schemes.

%The proposed algorithm adaptively partitions the model at two partitioning layers into three sub-models in a accuracy-aware manner and jointly determines their execution and aggregation across clients, enabling efficient parallel training of the sub-models.
%By explicitly accounting for model accuracy, computation and communication trade-offs, the algorithm significantly reduces training delay by 20\% and communication overhead by 50\% while improving model accuracy up to 3\%.
%Its performance is evaluated and compared against state-of-the-art SFL schemes through numerical experiments.
An important direction for future work is to further investigate the role of aggregators, particularly by studying the trade-offs between the number of aggregators and system performance metrics, such as accuracy, delay and overhead.
Another interesting direction would be to study the formulated problem from the perspective of the online optimization framework that makes decisions as clients’ processing tasks arrive or as the system’s state (e.g., client availability, connectivity, etc.) changes.
Another promising direction is to model the offloading decision as a multi-objective optimization problem that jointly considers training delay and energy consumption, and to develop adaptive strategies that identify efficient trade-offs under limited client availability.
%Also, its execution time remains acceptable for both medium and large-scale SFL systems, while its performance remains close to the optimal solution